\documentclass[sigconf,nonacm]{acmart}

\AtBeginDocument{%
  }

\usepackage{amsmath}
\usepackage{multirow}
\usepackage{enumitem}
\usepackage{listings}
\begin{document}

\title[LLM-Guided Cluster Lifecycle Management]{LLM-Guided Cluster Lifecycle
  Management: Deciding When and How to Evolve Behavioral Clusters}

\author{Himanshu Kulkarni}
\affiliation{%
  \institution{Amazon Web Services}
  \city{Seattle}
  \state{WA}
  \country{USA}}
\email{himankul@amazon.com}

\author{Siddharth Krishnamurthy}
\affiliation{%
  \institution{Amazon Web Services}
  \city{Seattle}
  \state{WA}
  \country{USA}}
\email{siddhakr@amazon.com}

\author{Jianhong Xin}
\affiliation{%
  \institution{Amazon Web Services}
  \city{Herndon}
  \state{VA}
  \country{USA}}
\email{xijianho@amazon.com}

\renewcommand{\shortauthors}{Kulkarni et al.}

\begin{abstract}
Behavioral clusters learned over user or item interactions drift as behavior evolves. Some fragment
into distinct sub-behaviors, some converge, some disappear, and new patterns emerge. Two paradigms
dominate the maintenance of such evolving clusterings. \emph{Threshold triggers} fire a pre-specified
action when a hardcoded statistic crosses a cutoff. \emph{Refit-and-track} (MONIC-style) re-clusters every
window and reads transitions off the alignment to the previous window. Both hardwire the \emph{decision}
into a threshold or a re-clustering rule. We study an alternative in which a large language model (LLM)
makes the lifecycle decision \emph{in place}. Given a threshold-free, whole-population view of the current
cluster set (statistical health, cluster labels, representative member texts, and residual mass), the LLM
decides for each cluster whether and how to act (\textsc{split}, \textsc{merge}, \textsc{retire},
\textsc{create}, or \textsc{no\_action}) without re-clustering. Which signal channel the LLM should use
depends on how informative the cluster labels are, so we fix a simple \emph{richness-conditioned modality
policy} before any window-level decision is made. We use the statistical channel when label distinctiveness
is low and fuse statistical$+$semantic when it is high, keyed on a label statistic computed in advance
and not picked after reading the results. On three public datasets spanning label richness (YooChoose
$0.17$, MOOC $0.50$, News-Category $0.70$), the policy-selected LLM mode obtains the highest F1 on all
three benchmarks, using far fewer interventions and showing stronger restraint than tuned threshold
heuristics and a MONIC steelman. All three head-to-head improvements are statistically significant (paired
Wilcoxon $p\le0.003$, bootstrap $95\%$ CIs excluding zero). Decomposing F1 shows where the gain comes from:
the LLM wins mainly on \emph{precision} (fewer false interventions), not on recall. Across our low-,
medium-, and high-richness benchmarks the useful modality shifts from statistics to semantics/fusion.
Statistics dominate when labels are impoverished (semantics \emph{hurt} on YooChoose), fusion is roughly
neutral at moderate richness (MOOC), and the semantic channel is required when labels are rich (on News the
statistical channel alone almost never acts and never produces a true positive). The same base prompt
template and the same fixed richness policy are used across all three datasets, whereas the
tuned baseline collapses when moved between them. We treat this shift as a useful practitioner diagnostic
observed on three datasets, not a validated law, and we specify the evaluation protocol (strict
direction-agnostic action matching, a restraint metric, and leakage-free as-of-window quality) in full
in the appendix.
\end{abstract}



\maketitle

\section{Introduction}
Clustering is rarely a one-shot operation in production systems. A clustering fit once is deployed and
then \emph{maintained} as new data arrives. Over time the population shifts. A broad cluster reveals two
sub-behaviors, two near-duplicate clusters converge, a cluster loses its members, or a new behavior
appears that no existing cluster explains. Maintaining the clustering means repeatedly answering a
decision problem. For each cluster, should we act, and if so which of \textsc{split}, \textsc{merge},
\textsc{retire}, \textsc{create}, or leave it alone?

Two families of methods answer this today. \textbf{Threshold triggers} monitor a statistic (Davies--Bouldin
index, cohesion $z$-score, centroid similarity) and fire a pre-specified action when it crosses a tuned
cutoff \cite{pattnayak2025lifecycle}. \textbf{Refit-and-track} methods, exemplified by MONIC
\cite{spiliopoulou2006monic}, re-cluster each window from scratch and infer transitions from how the new
clusters align to the previous ones. In both, the \emph{decision} is hardwired into a numeric threshold or
a re-clustering-plus-alignment rule, and both require per-dataset tuning (of thresholds, or of the
alignment floor $\tau$).

We study a third option: place a large language model (LLM) \emph{at the decision point}. Each window,
the LLM is shown a threshold-free, whole-population description of the current clusters (statistical health
signals, human-readable cluster labels, representative member texts, and the residual/unexplained mass),
and it decides, for every cluster, whether and how to intervene. It does not re-cluster. It evolves the
same persistent cluster registry in place. This reframes lifecycle maintenance from
\emph{trigger-then-execute} or \emph{refit-then-read} to \emph{reason-then-decide}.

\paragraph{Contributions.}
\begin{enumerate}[leftmargin=1.4em,itemsep=2pt]
  \item \textbf{A decision-layer formulation of cluster-lifecycle maintenance} (\S\ref{sec:method}): an LLM
        arbiter that decides in place over the whole cluster population, with no numeric cutoffs in the
        decision itself (``threshold-free'' is scoped to the decision layer, not the surrounding pipeline).
  \item \textbf{A strict, leakage-free evaluation protocol for lifecycle decisions} (\S\ref{sec:setup}):
        direction-agnostic \textsc{merge} matching that requires the correct pair, \textsc{create} matching
        that requires the created cluster to actually capture the intended population, a \emph{restraint}
        metric for whether stable clusters are left alone, and as-of-window quality scoring that removes
        time-travel leakage. The full protocol is specified in the appendix
        (Appendices~\ref{app:injection} and~\ref{app:hyper}).
  \item \textbf{Empirical evidence across three richness regimes} (\S\ref{sec:results}): on low-, medium-,
        and high-richness public benchmarks, the policy-selected LLM mode obtains the highest F1 on all
        three, with paired significance tests, and it does so with far fewer interventions. Decomposing F1
        shows the gain is concentrated in \emph{precision} rather than recall.
  \item \textbf{An observed modality/richness pattern} (\S\ref{sec:results}): across these three datasets
        the useful channel shifts from statistics to semantics/fusion as label distinctiveness rises. We
        offer this as a practical diagnostic for when to fuse, and a default boundary for acting on it, not
        as a validated law (\S\ref{sec:limitations}).
\end{enumerate}

We are explicit about what we do \emph{not} claim. The fused ``combined'' channel does not universally win
(on YooChoose it is well below stats-only), the richness boundary is a practical default rather than a
calibrated threshold, and three datasets cannot establish a general law. The defensible statement is that
\emph{LLM-based lifecycle arbitration improves decision quality per intervention, restraint, and
cross-dataset transfer}, with the useful modality tracking semantic richness.

\section{Related Work}
\paragraph{Evolving / streaming clustering.} Monitoring cluster transitions over time is classical.
MONIC \cite{spiliopoulou2006monic} formalizes external transitions (survival, split, absorption,
emergence, disappearance) by matching successive clusterings. We adopt MONIC's transition vocabulary and
use a modern refit-and-track implementation as our strongest baseline. Crucially, the
\textsc{split}/\textsc{merge}/\textsc{retire}/\textsc{create} vocabulary is \emph{not} claimed as our
novelty. It is MONIC's, and our contribution is the decision mechanism.

\paragraph{LLMs for clustering.} LLMs have been used to guide or perform clustering
\cite{zhang2023clusterllm,viswanathan2024fewshot,wan2024tntllm} and, in multi-agent form, to make
merge/split decisions \cite{lowe2026agentic}. These are largely \emph{static} (no temporal lifecycle) and
do not feed cluster-level health statistics to the model.

\paragraph{LLM-as-judge over statistical signals.} The closest pattern (statistical trigger
$\rightarrow$ LLM arbiter $\rightarrow$ streaming loop) appears for anomaly \emph{points} and drift
events \cite{devireddy2025calm,senol2025dkllm} and for self-healing model adaptation
\cite{rauba2024selfhealing}. Our decision object is different: it is the \emph{structure of a cluster
set}, the action space is structural, and the signals are cluster-level (cohesion-$z$,
split-improvement ratio, centroid similarity, membership trend, residual mass).

\paragraph{Trigger-then-execute vs.\ decide-in-place.} The most direct point of contrast is
lifecycle-aware clustering that uses a statistical trigger to decide \emph{when} and an LLM only to
\emph{execute} \cite{pattnayak2025lifecycle} (with a single action type). We move the decision itself to
the LLM, cover all five actions, and fuse statistical and semantic signals for the decision. To our
knowledge no prior work places an LLM at the cluster-lifecycle decision point with fused
statistical$+$semantic signals in a temporal setting.

\section{Problem Formulation}
Let a base clustering with $K$ clusters be fit at window $0$. At each subsequent window
$w=1,\dots,W$ a new batch of items arrives with feature vectors $X_w$. A \emph{maintenance method}
maintains a registry of persistent cluster identities and, at each window, emits a (possibly empty) set
of lifecycle actions drawn from \{\textsc{split}, \textsc{merge}, \textsc{retire}, \textsc{create},
\textsc{no\_action}\}. We evaluate against injected ground-truth events (\S\ref{sec:setup}): each event
specifies an action, a target cluster (or pair), and the window at which it should be detected. We do
\emph{not} evaluate proactive early detection. We ask, once drift has materialized, whether the method
takes the right action on the right cluster, and whether it avoids acting on clusters that did not change.

\section{Method}\label{sec:method}
\begin{figure*}[t]\centering
\includegraphics[width=0.95\linewidth]{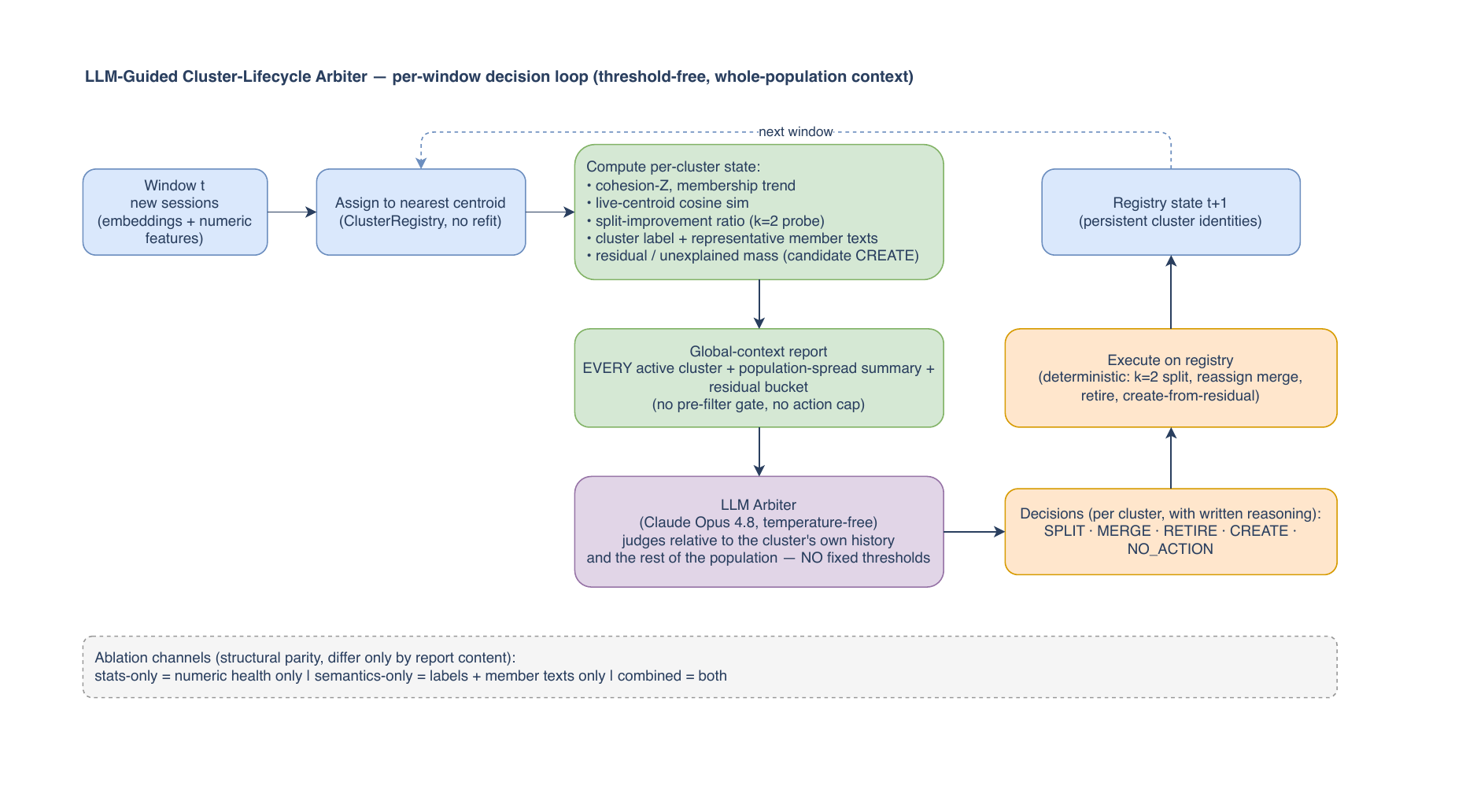}%
\caption{The per-window LLM-arbiter loop. New items are assigned to the current
registry (no refit). A threshold-free, whole-population report is built (statistical health $+$ labels $+$
representative member texts $+$ residual mass). The LLM decides all actions in place with written
rationale, and decisions are executed deterministically on the registry.}\Description{Flow diagram of the
per-window arbiter loop: new items are assigned to the existing cluster registry with no re-clustering; a
threshold-free whole-population report of statistical health, labels, representative member texts, and
residual mass is built; a single LLM decides an action for every cluster with a written rationale; and the
decisions are executed deterministically on the registry.}\label{fig:arch}\end{figure*}

\paragraph{Cluster registry.} We maintain a registry of persistent cluster identities with, per cluster,
its reference centroid, membership over windows, and a cohesion history (mean distance of members to the
centroid, higher is worse). Each window, new items are assigned to the nearest active centroid. No
re-clustering occurs. Actions mutate the registry deterministically: \textsc{split} via a $k{=}2$
sub-clustering of the members, \textsc{merge} by reassignment, \textsc{retire} by deactivation,
\textsc{create} by promoting a coherent residual mass to a new cluster.

\paragraph{Whole-population report.} For every active cluster we surface: cohesion and its $z$-score
relative to the cluster's own history, membership trend, live-centroid cosine similarity to neighbors, a
$k{=}2$ split-improvement ratio, the cluster's label, and a sample of representative member texts. We also
surface a \emph{residual bucket}: items poorly explained by any current centroid (a scale-free
mean$+k\sigma$ cut), with representative texts, the candidate for \textsc{create}. Critically the report
contains \emph{no} decision thresholds. The LLM is instructed to judge each cluster relative to its own
history and to the rest of the population. There is no pre-filter gate (every cluster is shown) and no cap
on the number of actions per window. An arbitrary cap would itself be a data-dependent threshold.

\emph{Scope of ``threshold-free.''} We use the term for the \textbf{decision layer}. The LLM is given no
numeric cutoffs and applies none in choosing actions. The surrounding pipeline retains a small number of
benign \emph{operational} heuristics that do not decide any action. Residual surfacing uses a scale-free
mean$+k\sigma$ cut ($k{=}1.5$) to \emph{populate} the candidate-\textsc{create} section, and executing a
\textsc{create} selects the orphan mass by a distance percentile. These affect what the LLM \emph{sees} and
how an approved action is \emph{carried out}, not whether or which action is taken. Throughout, ``threshold-free''
should be read as ``threshold-free decision layer,'' not as a claim that the entire system is free of
constants.

\paragraph{Arbiter.} The report is passed to a single LLM prompt (Claude Opus~4.8, which exposes no
temperature control, so the arbiter is near-deterministic) that returns, per cluster, an action and a
free-text rationale.
The \emph{same base prompt template is used on all datasets}, with no per-dataset wording, threshold, or
tuning. To be precise about what does vary: the richness policy selects \emph{which} of the three
structurally-parallel channel prompts is used (stats-only or combined), so the transferable object is the
template plus the fixed policy, not one identical input modality everywhere.

\paragraph{Ablation channels (structural parity).} To isolate what the LLM uses, three modes differ
\emph{only} in report content: \textbf{stats-only} (numeric health, no labels/texts),
\textbf{semantics-only} (labels $+$ member texts, no numbers), and \textbf{combined} (both). The combined
prompt is instructed to use either channel (a clear signal in one justifies action if the other does not
contradict), so it is not an AND-gate.

\paragraph{Richness-conditioned modality policy.}\label{par:policy}
Because the useful channel depends on how informative the cluster labels are, we avoid reporting a
per-dataset ``best-of-three.'' Instead we fix a policy $\pi$ keyed on a cheap label statistic, the mean
pairwise cosine distinctiveness $r$ of the cluster labels, computable before any window is processed:
\[
\pi(r)=\begin{cases}\textbf{stats-only} & r \le \tau_r\\[2pt] \textbf{combined} & r > \tau_r\end{cases}
\qquad \tau_r = 0.6 .
\]
The intuition is that impoverished labels ($r$ small) carry no discriminative semantic signal, so the
statistical channel should decide. Rich labels ($r$ large) make the semantic channel necessary, so both
should be fused. This is the single method we advocate. The three modes are its components and an ablation,
and we report per-mode numbers for transparency.

We are deliberate about what ``fixed'' means here. The \emph{form} of the policy (statistics at low
richness, fusion at high richness) follows from the mechanism, and $r$ is computed before any window-level
decision, so no per-window or per-seed tuning occurs. The boundary value $\tau_r = 0.6$, however, is a
default chosen with these three richness values in view ($0.17$, $0.50$, $0.70$), and we make no claim that
it is externally calibrated. Any $\tau_r \in (0.50, 0.695)$, whose upper bound is News's unrounded richness
(shown as $0.70$ elsewhere), produces the same three selections, so the
result is insensitive to the exact value within that interval, but localizing the boundary needs datasets
between those points. We treat its calibration as future work (\S\ref{sec:limitations}).

\section{Baselines}\label{sec:baselines}
\begin{itemize}[leftmargin=1.4em,itemsep=2pt]
  \item \textbf{No-Evolution} (frozen registry) and \textbf{Full-Retrain} (re-fits each window but records
        no decisions): lower-bound controls.
  \item \textbf{Statistical Triggers}: naive cutoffs (cohesion-$z\ge2$, similarity$>0.92$, empty for two
        windows, ${>}10\%$ orphans).
  \item \textbf{Smart Triggers}: a strong, hand-\emph{tuned} threshold heuristic with convergence checks
        and guards. \emph{We are explicit that this is a tuned-threshold baseline, not a clean
        ``same-scaffold-minus-LLM'' ablation}. Its thresholds are calibrated on MOOC and do not transfer.
  \item \textbf{MONIC} (steelman) \cite{spiliopoulou2006monic}: refit-and-track with per-window
        silhouette-selected $k$, centroid-cosine alignment ($\tau$ calibrated per dataset), and transition
        read-off. That MONIC needs a per-dataset $\tau$ while the arbiter uses one prompt is itself part
        of the transfer argument.
\end{itemize}

\section{Experimental Setup}\label{sec:setup}
\paragraph{Datasets and richness.} We use three public datasets chosen to span \emph{semantic label
richness} (mean pairwise cosine distinctiveness of cluster labels): \textbf{YooChoose}
(RecSys'15 clicks) \cite{yoochoose2015} at $0.17$, \textbf{MOOC} (SNAP user actions)
\cite{mooc2019snap} at $0.50$, and \textbf{News-Category} (HuffPost headlines, CC BY 4.0)
\cite{misra2022news} at $0.70$. Table~\ref{tab:datasets} summarizes them. All use MiniLM
embeddings \cite{reimers2019sbert}. MOOC/YooChoose use a $387$-d composite (embedding$\times0.8 \oplus$
scaled numeric features$\times0.2$), News uses L2-normalized $384$-d embeddings.

\begin{table*}[t]\centering\small
\caption{Public datasets. All: 12 windows, 15 seeds, 7 injected events (2 \textsc{split}, 2 \textsc{merge},
2 \textsc{retire}, 1 \textsc{create}).}\label{tab:datasets}
\begin{tabular}{lccccl}
\toprule
Dataset & Richness & $K$ & Items/seed & Space & Injection \\
\midrule
YooChoose & 0.17 & 10 & 35{,}000 & 387-d composite & vector perturbation \\
MOOC      & 0.50 & 10 & 33{,}591 & 387-d composite & vector perturbation \\
News      & 0.70 & 15 & 46{,}606 & 384-d L2-norm    & label composition \\
\bottomrule
\end{tabular}
\end{table*}

\paragraph{Drift injection.} MOOC/YooChoose inject drift by perturbing feature vectors (progressive
sub-population shift for \textsc{split}, absorption for \textsc{merge}, scatter-to-nearest for
\textsc{retire}, and a novel coherent far mass for \textsc{create}). News injects by \emph{curating which real
articles appear in which window} (real embeddings untouched): within-category \textsc{split} along a
category's own $k{=}2$ axis, near-duplicate category \textsc{merge}, low-volume category \textsc{retire},
and a held-out category \textsc{create}. Each seed also defines untouched \emph{control} clusters for the
restraint metric.

\paragraph{Evaluation metrics.} A decision counts as a true positive only under \emph{strict} matching
within $\pm2$ windows. The tolerance reflects that injected drift spreads across several windows, so a
correct decision may legitimately land a window early or late. We report precision (not recall alone)
throughout, so a method that merely saturates windows to overlap the true one gains no advantage. Appendix~\ref{sec:supp-tol} re-scores every method
at $\pm1$, $\pm2$, and $\pm3$ windows and finds the winner unchanged on two of three datasets at all
tolerances (and on the third at $\pm2$ and $\pm3$). \textsc{split} and \textsc{retire} require the correct target. \textsc{merge} is
matched \emph{direction-agnostically} but requires \emph{both} clusters of the fused pair to match the
ground-truth pair (merging a cluster with the wrong partner is not credited). \textsc{create} is credited
only if $\ge50\%$ of the created cluster's members belong to the intended emergent population
(\emph{strict CREATE}). From these we report precision/recall/\textbf{F1}, per-action F1, and
\emph{intervention economy} (actions per run). We add two metrics beyond accuracy:
\begin{itemize}[leftmargin=1.4em,itemsep=1pt]
  \item \textbf{Restraint}: the fraction of ground-truth \textsc{no\_action} control clusters left
        untouched (never split/merged/retired, nor named as a merge target). This is the clearest contrast
        with refit-and-track, which re-derives the partition every window and churns stable clusters. It
        is meaningful only paired with recall (a do-nothing method scores $1.0$ trivially).
  \item \textbf{As-of-window cluster quality} (silhouette): each window is scored using only clusters that
        existed as of that window (reconstructed from created/retired windows), removing the time-travel
        leakage of scoring against the final registry. Reported as descriptive only.
\end{itemize}

\paragraph{Protocol.} We run $15$ seeds per dataset across all $8$ methods, and report mean F1 with $95\%$ CIs and,
for the best-LLM-vs-best-baseline contrast per dataset, a one-sided paired Wilcoxon signed-rank test and a
$10{,}000$-sample paired bootstrap $95\%$ CI on the F1 delta. Arbiter LLM is Claude Opus~4.8 with a single
prompt across datasets.

\section{Results}\label{sec:results}
\paragraph{How to read these numbers.} Absolute F1 sits between $0.2$ and $0.35$ for the best methods, which
is low in isolation and worth explaining before the tables. Three properties of the benchmark suppress it by
design. First, the scorer is strict: a \textsc{merge} that fuses a ground-truth cluster with the
\emph{wrong} partner earns zero credit, and a \textsc{create} that fires in the right window but scoops an
incoherent mass is denied unless the new cluster is majority the intended population. Second, the injected
events are deliberately subtle, so that detecting them requires reading sustained
change instead of a single spike. On News the within-category \textsc{split} is injected at its natural
$\sim$25\% minority share and is not detected at usable precision by any method (\S\ref{par:newssplit}),
which alone caps the effective ceiling near 6 of 7 events. Third, recall is bounded by these 7 events per seed, so a single
missed event costs $\approx$0.14 recall. The quantity that carries meaning here is therefore the
\emph{relative} comparison under an identical scorer, which is why we report paired per-seed tests
(Table~\ref{tab:sig}) instead of point estimates.

\begin{table*}[t]\centering\small
\caption{Main results over 15 seeds (mean $\pm$ 95\% CI on every column). Best F1 per dataset in
\textbf{bold}, and $\pi$ marks the mode chosen by the richness policy. The Actions column (intervention
economy) is also bolded on the $\pi$ row, showing the policy-selected mode reaches the best F1 with among the
fewest interventions of any detecting method. Reading precision and recall separately
shows the LLM's advantage is concentrated in precision: on MOOC the tuned baseline has \emph{higher} recall
($0.381$ vs.\ $0.324$) but roughly half the precision.}\label{tab:main}
\begin{tabular}{llccccc}
\toprule
Dataset & Method & Precision & Recall & F1 & Actions & Restraint \\
\midrule
\multirow{6}{*}{\rotatebox{90}{YooChoose}}
 & Stat-Triggers        & $0.030\pm0.005$ & $0.171\pm0.030$ & $0.051\pm0.008$ & $39.6\pm0.7$ & $0.00\pm0.00$ \\
 & Smart-Triggers       & $0.177\pm0.028$ & $0.305\pm0.046$ & $0.223\pm0.035$ & $12.1\pm0.7$ & $0.17\pm0.12$ \\
 & MONIC                & $0.071\pm0.001$ & $0.143\pm0.000$ & $0.095\pm0.001$ & $14.1\pm0.3$ & $0.00\pm0.00$ \\
 & LLM stats-only $\pi$ & $0.280\pm0.049$ & $0.429\pm0.055$ & $\mathbf{0.336\pm0.050}$ & $\mathbf{11.2\pm1.1}$ & $0.97\pm0.07$ \\
 & LLM semantics-only   & $0.058\pm0.003$ & $0.143\pm0.000$ & $0.083\pm0.003$ & $17.3\pm0.8$ & $0.00\pm0.00$ \\
 & LLM combined         & $0.153\pm0.040$ & $0.210\pm0.054$ & $0.176\pm0.044$ & $\phantom{0}9.4\pm1.0$ & $0.97\pm0.07$ \\
\midrule
\multirow{6}{*}{\rotatebox{90}{MOOC}}
 & Stat-Triggers        & $0.036\pm0.001$ & $0.143\pm0.000$ & $0.058\pm0.001$ & $27.5\pm0.8$ & $0.00\pm0.00$ \\
 & Smart-Triggers       & $0.214\pm0.035$ & $0.381\pm0.065$ & $0.272\pm0.043$ & $12.6\pm1.0$ & $0.37\pm0.12$ \\
 & MONIC                & $0.051\pm0.004$ & $0.162\pm0.025$ & $0.077\pm0.007$ & $22.1\pm1.6$ & $0.00\pm0.00$ \\
 & LLM stats-only $\pi$ & $0.389\pm0.043$ & $0.324\pm0.033$ & $\mathbf{0.348\pm0.028}$ & $\mathbf{\phantom{0}6.1\pm0.9}$ & $0.53\pm0.07$ \\
 & LLM semantics-only   & $0.242\pm0.054$ & $0.333\pm0.052$ & $0.278\pm0.052$ & $10.3\pm1.0$ & $0.33\pm0.12$ \\
 & LLM combined         & $0.347\pm0.047$ & $0.257\pm0.041$ & $0.286\pm0.036$ & $\phantom{0}5.5\pm1.0$ & $0.60\pm0.11$ \\
\midrule
\multirow{6}{*}{\rotatebox{90}{News}}
 & Stat-Triggers        & $0.053\pm0.010$ & $0.219\pm0.037$ & $0.086\pm0.016$ & $29.5\pm2.1$ & $0.25\pm0.04$ \\
 & Smart-Triggers       & $0.000\pm0.000$ & $0.000\pm0.000$ & $0.000\pm0.000$ & $\phantom{0}0.0\pm0.0$ & $1.00\pm0.00$ \\
 & MONIC                & $0.081\pm0.027$ & $0.219\pm0.066$ & $0.116\pm0.037$ & $20.2\pm2.6$ & $0.76\pm0.07$ \\
 & LLM stats-only       & $0.000\pm0.000$ & $0.000\pm0.000$ & $0.000\pm0.000$ & $\phantom{0}0.2\pm0.3$ & $1.00\pm0.00$ \\
 & LLM semantics-only   & $0.144\pm0.031$ & $0.171\pm0.030$ & $0.155\pm0.030$ & $\phantom{0}8.7\pm0.9$ & $0.99\pm0.02$ \\
 & LLM combined $\pi$   & $0.506\pm0.136$ & $0.152\pm0.019$ & $\mathbf{0.228\pm0.032}$ & $\mathbf{\phantom{0}2.5\pm0.4}$ & $1.00\pm0.00$ \\
\bottomrule
\end{tabular}
\end{table*}

\paragraph{The richness policy selects the strongest LLM channel on all three benchmarks.}
The policy $\pi$ (\S\ref{par:policy}) selects stats-only for YooChoose ($r{=}0.17$) and MOOC ($r{=}0.50$)
and combined for News ($r{=}0.70$). Applied from the label statistic alone, it lands on the highest-F1 mode
in all three cases, so the headline is a fixed rule rather than a per-dataset selection (the only
dataset-specific input is $r$, measured before any window-level decision).
Table~\ref{tab:main} shows the policy-selected mode beats the best baseline by
$+0.075$ (MOOC), $+0.111$ (News), $+0.113$ (YooChoose) in F1. All three contrasts are significant with
bootstrap CIs excluding zero (Table~\ref{tab:sig}). The LLM also intervenes far less
(2.5--11 actions vs.\ 12--40 for triggers/MONIC) and leaves stable clusters alone
(restraint $0.53$--$1.00$, against $0.00$ for MONIC and naive triggers on YooChoose and MOOC, where they
disturb \emph{every} control cluster, and $0.76$ and $0.25$ respectively on News). A label ablation confirms this
modality pattern is causal, not just correlational. Destroying the cluster labels collapses News-combined
to the stats-only floor ($0.228\to0.000$), while neutralizing YooChoose's uninformative labels slightly
\emph{improves} it, indicating those labels were mildly misleading (Appendix~\ref{app:robustness}, Table~\ref{tab:ablation}).

\paragraph{The gain is precision, not recall.} Decomposing F1 (Table~\ref{tab:main}) locates the advantage
precisely, and the honest version is narrower than ``the LLM detects more.'' On MOOC the tuned
Smart-Triggers baseline achieves \emph{higher} recall than the policy-selected LLM ($0.381$ vs.\ $0.324$)
while scoring far lower precision ($0.214$ vs.\ $0.389$). It finds marginally more events but pays about
$10$ false interventions per run against the LLM's $3.8$. On News the effect is starker: combined reaches
precision $0.506$ at recall $0.152$, with $1.4$ false positives per run, where MONIC spends $18.7$ false
positives to reach comparable recall ($0.219$). Only on YooChoose does the LLM lead on both axes
(precision $0.280$ vs.\ $0.177$, recall $0.429$ vs.\ $0.305$). The practical claim is therefore about the
\emph{cost of maintenance}: for a comparable number of correct decisions, the arbiter issues a small
fraction of the incorrect ones.

\begin{table*}[t]\centering\small
\caption{Focal significance: best LLM vs.\ best baseline per dataset (paired, 15 seeds). $\Delta$F1 with
$10^4$-sample bootstrap 95\% CI and one-sided paired Wilcoxon $p$.}\label{tab:sig}
\begin{tabular}{lllccc}
\toprule
Dataset & LLM & Baseline & $\Delta$F1 & 95\% CI & $p$ \\
\midrule
MOOC      & stats-only & Smart-Trig. & $+0.075$ & $[0.038,0.117]$ & 0.0008 \\
News      & combined   & MONIC       & $+0.111$ & $[0.068,0.156]$ & 0.0009 \\
YooChoose & stats-only & Smart-Trig. & $+0.113$ & $[0.055,0.170]$ & 0.0027 \\
\bottomrule
\end{tabular}
\end{table*}

\begin{figure}[t]\centering
\includegraphics[width=0.72\linewidth]{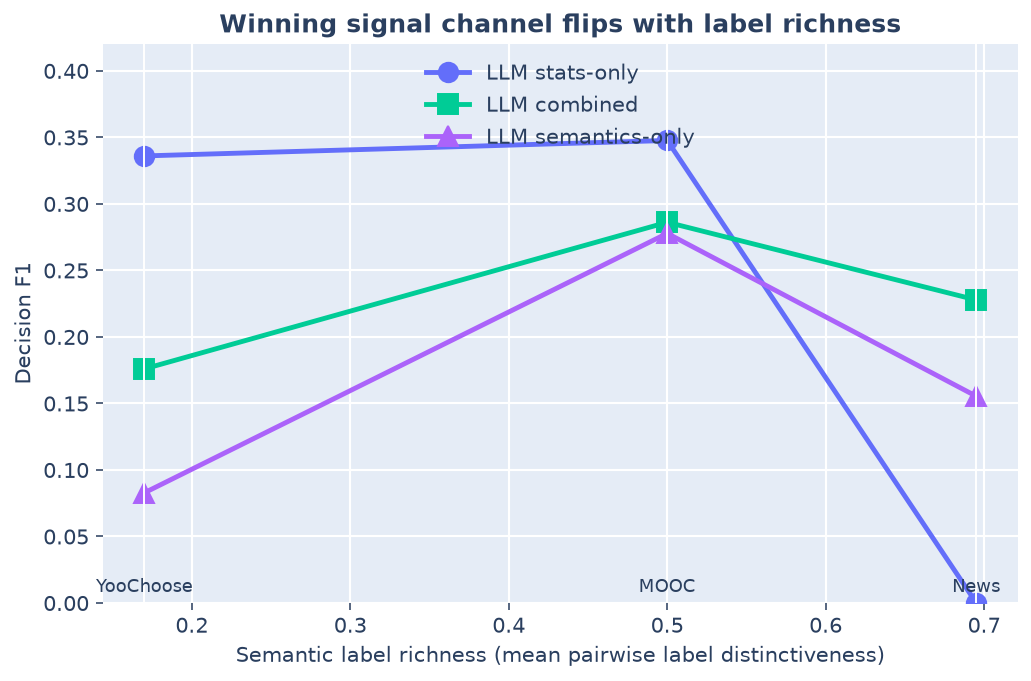}
\caption{Semantic-richness spectrum: LLM channel F1 vs.\ label richness across the three datasets.}
\Description{Line plot of decision F1 against cluster-label richness for the three LLM channels across the
datasets. The statistics-only channel is strongest at low richness while the combined channel becomes
strongest at high richness, so the winning channel shifts from statistics to fusion as richness rises.}
\label{fig:spectrum}
\end{figure}

\paragraph{Across low-, medium-, and high-richness benchmarks the useful modality shifts.}
Figure~\ref{fig:spectrum} shows the shift. On \textbf{YooChoose} (0.17) stats-only wins ($0.336$) and the
semantic channel \emph{hurts} (semantics-only $0.083$, dragging combined to $0.176$). On \textbf{MOOC}
(0.50) stats-only is best ($0.348$) with fusion roughly neutral. On \textbf{News} (0.70) the semantic
channel is required: combined is best ($0.228$) while stats-only almost never acts and never produces a true
positive ($0.000$ F1 at $0.2\pm0.3$ actions per run).
Ordered by richness, the semantic channel moves from harmful to neutral to necessary. We read this as an
observed pattern over three datasets, sufficient to act on as a diagnostic (\emph{choose the modality by
label richness}) but not as a validated law: three points cannot establish monotonicity, and the mechanism
we propose for it (semantic discrimination requires discriminative labels) is an interpretation rather than
a tested hypothesis. It is, however, enough to refute the stronger claim that fusion is universally robust,
since YooChoose is a direct counterexample.

\paragraph{Per-action view.}
Table~\ref{tab:peraction} refines this. Contrary to a blanket ``CREATE is weak,''
\textsc{create} F1 reaches $0.49$ (MOOC, semantics) and $0.31$ (YooChoose, semantics). It is $0$ on News
only because strict scoring correctly denies the low-purity emergent mass there. \textsc{split} and
\textsc{retire} are strong on MOOC/YooChoose. \textsc{merge} carries News (the near-duplicate-category
signal is semantic by construction). \textsc{split} on News is a null for all three LLM modes, and we treat
it as undetected by \emph{any} method at usable precision (\S\ref{par:newssplit}).

\begin{table*}[t]\centering\small
\caption{Per-action F1 for the LLM modes (strict scorer, 15 seeds). ``--'' $=0$.}\label{tab:peraction}
\begin{tabular}{lllcccc}
\toprule
Dataset & Mode & & \textsc{split} & \textsc{merge} & \textsc{retire} & \textsc{create} \\
\midrule
\multirow{3}{*}{YooChoose}
 & stats-only     && 0.46 & --   & 0.44 & --   \\
 & semantics-only && --   & --   & --   & 0.31 \\
 & combined       && 0.38 & 0.04 & 0.18 & --   \\
\midrule
\multirow{3}{*}{MOOC}
 & stats-only     && 0.48 & --   & 0.48 & --   \\
 & semantics-only && --   & 0.18 & 0.35 & 0.49 \\
 & combined       && 0.36 & --   & 0.54 & --   \\
\midrule
\multirow{3}{*}{News}
 & stats-only     && --   & --   & --   & --   \\
 & semantics-only && --   & 0.54 & --   & --   \\
 & combined       && --   & 0.49 & --   & --   \\
\bottomrule
\end{tabular}
\end{table*}

\paragraph{Is the LLM better on the action types that matter, or merely less noisy?}
Aggregate F1 cannot answer this, so Table~\ref{tab:peraction_base} puts the policy-selected LLM against the
best baseline on each dataset, action by action. The answer is mixed, and we report it as such. The LLM wins
the actions that depend on reading sustained membership decline or semantic identity: \textsc{retire} on
YooChoose ($0.439$ vs.\ $0.000$) and MOOC ($0.476$ vs.\ $0.394$), and \textsc{merge} on News ($0.485$ vs.\
$0.000$). The baselines win where a single geometric cue suffices: Smart-Triggers beats the LLM on
\textsc{split} on YooChoose ($0.566$ vs.\ $0.460$) and on \textsc{create} on MOOC ($0.210$ vs.\ $0.000$),
and MONIC scores non-zero on News \textsc{split}/\textsc{retire} ($0.182$/$0.175$) where the LLM scores zero.
So the LLM is not uniformly stronger per action. Its aggregate lead comes from being strong on the
harder, context-dependent actions while not paying the false-positive cost the baselines incur elsewhere.

\begin{table*}[t]\centering\small
\caption{Per-action F1: best baseline vs.\ the policy-selected LLM mode on each dataset (strict scorer, 15
seeds). Baselines are Smart-Triggers (YooChoose, MOOC) and MONIC (News), each the strongest baseline by
aggregate F1 on that dataset. Bold marks the winner of each row, and ``--'' $=0$.}\label{tab:peraction_base}
\begin{tabular}{llcccc}
\toprule
Dataset & Method & \textsc{split} & \textsc{merge} & \textsc{retire} & \textsc{create} \\
\midrule
\multirow{2}{*}{YooChoose}
 & Smart-Triggers (best baseline) & $\mathbf{0.566}$ & $\mathbf{0.060}$ & -- & -- \\
 & LLM stats-only ($\pi$)         & $0.460$ & -- & $\mathbf{0.439}$ & -- \\
\midrule
\multirow{2}{*}{MOOC}
 & Smart-Triggers (best baseline) & $0.443$ & -- & $0.394$ & $\mathbf{0.210}$ \\
 & LLM stats-only ($\pi$)         & $\mathbf{0.475}$ & -- & $\mathbf{0.476}$ & -- \\
\midrule
\multirow{2}{*}{News}
 & MONIC (best baseline)          & $\mathbf{0.182}$ & -- & $\mathbf{0.175}$ & -- \\
 & LLM combined ($\pi$)           & -- & $\mathbf{0.485}$ & -- & -- \\
\bottomrule
\end{tabular}
\end{table*}

\paragraph{Why the baselines' non-zero News \textsc{split}/\textsc{retire} is not detection.}\label{par:newssplit}
Table~\ref{tab:peraction_base} shows MONIC scoring $0.182$ on News \textsc{split} and $0.175$ on
\textsc{retire} where every LLM mode scores zero, which appears to contradict our statement that the News
\textsc{split} is undetected. It does not, and the mechanism is worth stating because it recurs whenever a
method is allowed to act freely. Those non-zero values come from near-indiscriminate firing rather than from
identifying the event. Summed over 15 seeds, MONIC's News \textsc{split} record is 5 true positives against
20 false positives (precision $0.200$), and its \textsc{retire} record is 18 true positives against
\emph{158} false positives (precision $0.102$). Statistical Triggers is the extreme case: it attains recall
$0.767$ on News \textsc{split}, nominally the best of any method, by emitting 318 false positives to collect
23 true ones (precision $0.067$). Under a $\pm2$-window matching rule, a method that fires on most clusters
in most windows will overlap the true event window by construction, so recall alone cannot distinguish
detection from saturation. We therefore describe the News \textsc{split} as undetected \emph{at usable
precision} by any method, rather than as strictly zero for all methods, and we report precision alongside
recall throughout so this distinction is visible. The same caution applies to any future baseline that buys
recall with volume.

\paragraph{Error analysis.} The false-positive and false-negative structure differs by dataset in ways the
richness pattern predicts. On \textbf{YooChoose} ($r{=}0.17$) numeric drift is the only informative signal,
and stats-only exploits it (3.0 true positives, 8.2 false positives per run). The semantic channel misreads
near-identical labels and over-fires, producing the worst error profile of any method here (1.0 true
positives against 16.3 false positives), which is what drags fused combined below stats-only. On
\textbf{MOOC} ($r{=}0.50$) the two channels are close and their errors are complementary: semantics-only is
the only \emph{LLM} mode that detects \textsc{create} ($0.49$, against $0.210$ for the tuned baseline)
because the emergent behavior is describable, but it is
weaker on \textsc{split}, so fusing yields no net gain (combined $0.286$ vs.\ stats-only $0.348$) while
halving false positives relative to the tuned baseline. On \textbf{News} ($r{=}0.70$) the statistical
channel almost never acts and never lands a true positive ($0.2\pm0.3$ actions per run, precision and recall
both zero) because the
label-composition drift leaves the embedding geometry almost unchanged. \textsc{merge} is carried entirely
by reading near-duplicate category labels, while strict \textsc{create} and the natural-share
\textsc{split} remain undetected by every method. Two of the three dominant failure modes are therefore
properties of the benchmark, not of the arbiter. The News \textsc{split} is undetected at usable
precision by any method, and strict \textsc{create} correctly denies low-purity emergent mass.

\paragraph{Transfer.}
The tuned Smart-Triggers baseline scores $0.272$ on MOOC but $0.000$ on News (its MOOC-calibrated
thresholds never fire in the normalized News space). MONIC needs a per-dataset alignment floor $\tau$. The
LLM arbiter uses the \emph{same base prompt template and the same fixed richness policy} on all three and is
best throughout, evidence that moving the decision to a threshold-free reasoner improves portability. The
policy does switch channel between datasets, but it does so from a statistic computed in advance, without the
per-dataset threshold tuning the baselines rely on. On semantic drift the advantage
is categorical, not incremental. No threshold method detects the News split even when its thresholds
are \emph{oracle-tuned} to the evaluation labels (F1 $=0$, unattainable in deployment), whereas the arbiter
reaches $0.228$; a same-numeric-feature learned classifier likewise fails to transfer without per-dataset
supervision (Appendix~\ref{app:robustness}, Tables~\ref{tab:oracle} and~\ref{tab:samefeature}).

\paragraph{Economy and restraint, and why restraint is not a free win.}
The LLM modes combine high F1 with few actions, whereas naive
triggers over-fire (28--40 actions) and MONIC churns (14--22). Restraint (Table~\ref{tab:main}) is
near-perfect for the LLM on News and YooChoose ($0.97$--$1.00$) and clearly above the mechanical baselines on
MOOC ($0.53$--$0.60$ vs.\ $0.00$--$0.37$), with the naive triggers and MONIC disturbing \emph{every} control
cluster on both YooChoose and MOOC.

Restraint on its own is trivially gamed: a method that never acts scores $1.00$, which is precisely why
we run No-Evolution and Full-Retrain as lower-bound controls (\S\ref{sec:baselines}). Neither records a
lifecycle decision, so both have zero recall, and we omit them from the tables. The claim only means something \emph{paired with recall},
and that pairing is what the results support. On YooChoose the policy-selected LLM holds restraint $0.97$
while achieving the highest recall of any method ($0.429$ vs.\ $0.305$ for the tuned baseline). On MOOC it
holds restraint $0.53$--$0.60$ at recall within one standard error of the tuned baseline. On News combined
holds restraint $1.00$ at recall $0.152$, against MONIC's $0.219$ recall bought with restraint $0.76$ and
$18.7$ false positives per run. In other words the arbiter is not being credited for inaction: it leaves
stable clusters alone \emph{and} still detects a comparable or larger share of the injected events. The
practitioner reading is that restraint measures the cost side of maintenance, and it is only interpretable
next to the detection side.

\section{Discussion}
\paragraph{Where the decision lives matters.} Moving the decision from a threshold/refit rule to an LLM
reasoning over the full population yields better decisions per intervention, stronger restraint, and
one-prompt transfer. Figure~\ref{fig:paradigms} illustrates the paradigm contrast.

\paragraph{When to fuse.} The spectrum gives a concrete rule: with impoverished labels, use statistics
(semantics add noise and can hurt). With rich labels, the semantic channel is required. Fusion is
recommended only when labels carry information. It is \emph{not} robust to a harmful channel.

\section{Limitations}\label{sec:limitations}
We state these directly, because most of them bound the interpretation of the headline result.

\textbf{Dataset count.} Three public datasets span useful richness regimes ($0.17$, $0.50$, $0.70$) but
cannot establish a general richness law. The modality pattern is an observation over three points, and the
ordering could fail on datasets whose labels are rich but uninformative for lifecycle change specifically.

\textbf{Policy calibration.} $\tau_r = 0.6$ is a practical default, not a calibrated boundary. It was chosen
with these three richness values in view. Any value in $(0.50, 0.695)$ (the upper bound is News's unrounded
richness, shown as $0.70$ elsewhere) gives identical selections here, so the result is insensitive within
that band, but locating the true boundary requires datasets between $0.50$ and $0.70$.

\textbf{Injected drift.} The public evaluation uses code-injected events. This is what makes exact
ground truth and a strict scorer possible, but injected drift need not exhaust the ways real clusterings
decay, and the injection design inevitably encodes assumptions about what a \textsc{split} or \textsc{merge}
looks like.

\textbf{Unequal action difficulty.} Action types are not equally detectable under this protocol. News
\textsc{split} is undetected at usable precision by any method (\S\ref{par:newssplit}), strict
\textsc{create} is deliberately conservative, and
\textsc{merge} requires the correct partner. Aggregate F1 therefore mixes actions of very different
difficulty, which is why we report the per-action breakdown (Tables~\ref{tab:peraction},
\ref{tab:peraction_base}).

\textbf{LLM dependence.} All results use one frontier model and one prompt. We did not test cheaper or
smaller models, prompt paraphrases, or sensitivity to the report's wording and field order, so we cannot
separate ``an LLM at the decision point helps'' from ``this model with this prompt helps.''

\textbf{Baseline scope.} Smart-Triggers is a hand-tuned heuristic, not a learned non-LLM arbiter. We include
a learned classifier over the same \emph{numeric} report signals (Appendix~\ref{app:robustness}, Table~\ref{tab:samefeature}), which does not
match the LLM even given per-dataset supervision. A learned model consuming the semantic labels and member
texts in a comparable representation is still absent, so the architecture-versus-representation separation is
not total. We also do not exhaust per-dataset threshold tuning for the trigger family. Smart-Triggers is
calibrated on MOOC to test whether a plausible hand-engineered policy transfers, and MONIC receives
per-dataset alignment calibration, so the trigger family is represented at both fixed and tuned settings but
not at a per-dataset optimum. A trigger suite tuned to each dataset could raise baseline F1. Such tuning
requires labeled drift on every target dataset, which is supervision the single-prompt arbiter is never
given, so we treat a fully per-dataset-tuned trigger suite as an upper-bound comparison for future work.

\textbf{Cost accounting.} We report arbiter LLM \emph{calls} ($\approx$1 per window, $\approx$10 per
seed-run) instead of tokens, because the inference client we used did not expose per-call token usage for
this model. Absolute cost comparisons against cheaper baselines are therefore approximate. Cost scales with the
number of active clusters and the representative examples shown per cluster, not with the number of raw
items, because the arbiter reasons over a summarized cluster report. On
MOOC, whose policy mode is stats-only over 10 active clusters, the report averaged about 1,700 input tokens
per window under a chars/4 approximation, roughly 400 characters per cluster, with about one arbiter call
per window. Modes that add representative member texts per cluster produce proportionally larger reports.

\textbf{Operational constants.} The decision layer is threshold-free, but the surrounding pipeline is not.
Residual surfacing uses a mean$+1.5\sigma$ cut and \textsc{create} execution selects the orphan mass by a
distance percentile. These shape what the arbiter sees and how an approved action is carried out, and a
different choice could change which residual groups are ever considered.

\paragraph{Future work.} Several directions follow directly. The \emph{label-ablation} in Appendix~\ref{app:robustness} already establishes semantic causality in
the necessity and harm directions. Shuffling and genericizing the labels collapses the semantic gain on News
and shows YooChoose's impoverished labels are mildly misleading rather than useful. The one remaining arm is
the \emph{sufficiency} direction, enriching impoverished labels to test whether the semantic channel would
then help, which we leave to future work. Beyond
that, validate the richness boundary on more datasets, particularly between $r{=}0.50$ and $r{=}0.70$. Add
model and prompt robustness checks across model tiers, prompt paraphrases, and report field order, which
would separate ``the decision-point architecture helps'' from ``this particular prompt helps.'' The
same-numeric-feature learned arbiter is already reported (Table~\ref{tab:samefeature}). A learned arbiter over the
\emph{semantic} channel (labels and texts) would complete the architecture-versus-representation ablation.
Characterize sensitivity
to event severity by sweeping split magnitude, merge strength, retirement speed, and emergent mass, which
would map where each method starts to fail rather than only whether it fails. Run control-only streams with
no injected events, which stress-tests restraint and false-positive behavior directly and removes the
do-nothing confound noted above. Vary the operational constants (residual surfacing cut, \textsc{create}
percentile, split-execution settings) to confirm the threshold-free decision layer is robust to the
pipeline around it. Finally, evaluate natural (non-injected) drift with human adjudication, where no ground
truth exists but expert judgment can substitute, and re-run cost accounting on a usage-exposing client so
the intervention-economy claim covers computational cost and not only action counts.

\section{Conclusion}
Placing an LLM at the cluster-lifecycle decision point (deciding in place, with a threshold-free decision
layer, over the whole population) improves evolving-cluster maintenance over tuned threshold triggers and a
MONIC steelman on three public datasets, with paired significance and far fewer, more targeted
interventions, using a single transferable prompt. Decomposing the gain shows it is concentrated in
precision: for a comparable number of correct decisions, the arbiter issues a small fraction of the
incorrect ones. A fixed richness-conditioned policy selects the right channel on all three benchmarks, and
the modality shift it exploits gives practitioners a usable diagnostic for when fusing semantics is worth
it. To support auditing without claiming turnkey reproducibility, the appendix provides, for all three public
datasets, the arbiter prompt and its channel variants, the cluster-report schema with a worked example, the
preprocessing and drift-injection protocols, the seeds, the baseline thresholds and MONIC calibration, and
the strict-scorer rules. These are sufficient to audit the evaluation protocol and inspect every reported
table, significance test, and sensitivity analysis. The three public corpora are cited above and remain
available from their original sources.

\section*{Ethical considerations}
All three datasets are public research corpora used under their stated licenses (YooChoose from the
RecSys'15 challenge, the SNAP MOOC user-action log, and the CC BY 4.0 News-Category corpus), and all lifecycle
events are code-injected, so no personal data is collected or released. The method is a decision-support
layer above an existing clustering rather than an autonomous actor. Every action and every
\textsc{no\_action} carries a written rationale and is executed deterministically, so a maintainer can
audit and reverse a decision at the next window. Two risks deserve statement. First, a method that prefers
\textsc{no\_action} under weak evidence will describe a genuinely new or minority behavior later than an
aggressive baseline, so small sub-populations can stay under-represented until enough evidence accumulates.
Second, the arbiter reasons over model-generated cluster labels and member texts, so any bias in the
labeling model or in which behaviors are sampled propagates into its decisions, and high-volume behaviors
are described by richer evidence than rare ones. The richness-conditioned policy mitigates the first-order
version of this by declining the semantic channel where labels carry little signal instead of fusing it by
default. We recommend that deployments keep the human-auditable log and treat the arbiter as advisory to a
maintainer, not an unsupervised controller.

\bibliographystyle{ACM-Reference-Format}
\bibliography{refs}


\begin{thebibliography}{13}


\ifx \showCODEN    \undefined \def \showCODEN     #1{\unskip}     \fi
\ifx \showISBNx    \undefined \def \showISBNx     #1{\unskip}     \fi
\ifx \showISBNxiii \undefined \def \showISBNxiii  #1{\unskip}     \fi
\ifx \showISSN     \undefined \def \showISSN      #1{\unskip}     \fi
\ifx \showLCCN     \undefined \def \showLCCN      #1{\unskip}     \fi
\ifx \shownote     \undefined \def \shownote      #1{#1}          \fi
\ifx \showarticletitle \undefined \def \showarticletitle #1{#1}   \fi
\ifx \showURL      \undefined \def \showURL       {\relax}        \fi
\providecommand\bibfield[2]{#2}
\providecommand\bibinfo[2]{#2}
\providecommand\natexlab[1]{#1}
\providecommand\showeprint[2][]{arXiv:#2}

\bibitem[Devireddy and Huang(2025)]%
        {devireddy2025calm}
\bibfield{author}{\bibinfo{person}{Ashok Devireddy} {and}
  \bibinfo{person}{Shunping Huang}.} \bibinfo{year}{2025}\natexlab{}.
\newblock \bibinfo{title}{{CALM}: A Framework for Continuous, Adaptive, and
  {LLM}-Mediated Anomaly Detection in Time-Series Streams}.
\newblock
\shownote{arXiv:2508.21273}.
\newblock


\bibitem[L{\"o}we and Silcock(2026)]%
        {lowe2026agentic}
\bibfield{author}{\bibinfo{person}{Simon L{\"o}we} {and} \bibinfo{person}{Emily
  Silcock}.} \bibinfo{year}{2026}\natexlab{}.
\newblock \bibinfo{title}{Agentic Clustering: Controllable Text Taxonomies via
  Multi-Agent Refinement}.
\newblock
\shownote{arXiv:2606.01255}.
\newblock


\bibitem[Misra(2022)]%
        {misra2022news}
\bibfield{author}{\bibinfo{person}{Rishabh Misra}.}
  \bibinfo{year}{2022}\natexlab{}.
\newblock \bibinfo{title}{News Category Dataset}.
\newblock
\shownote{Kaggle / arXiv:2209.11429; CC BY 4.0 license}.
\newblock


\bibitem[Pattnayak et~al\mbox{.}(2025)]%
        {pattnayak2025lifecycle}
\bibfield{author}{\bibinfo{person}{Priyaranjan Pattnayak},
  \bibinfo{person}{Sanchari Chowdhuri}, \bibinfo{person}{Amit Agarwal}, {and}
  \bibinfo{person}{Hitesh~Laxmichand Patel}.} \bibinfo{year}{2025}\natexlab{}.
\newblock \showarticletitle{{LLM}-Guided Lifecycle-Aware Clustering of
  Multi-Turn Customer Support Conversations}. In
  \bibinfo{booktitle}{\emph{Proceedings of the Association for Computational
  Linguistics (AACL)}}.
\newblock
\shownote{arXiv:2601.04388; verified}.
\newblock


\bibitem[Rauba et~al\mbox{.}(2024)]%
        {rauba2024selfhealing}
\bibfield{author}{\bibinfo{person}{Paulius Rauba}, \bibinfo{person}{Nabeel
  Seedat}, \bibinfo{person}{Krzysztof Kacprzyk}, {and} \bibinfo{person}{Mihaela
  van~der Schaar}.} \bibinfo{year}{2024}\natexlab{}.
\newblock \showarticletitle{Self-Healing Machine Learning: A Framework for
  Autonomous Adaptation in Real-World Environments}. In
  \bibinfo{booktitle}{\emph{Advances in Neural Information Processing Systems
  (NeurIPS)}}.
\newblock
\shownote{arXiv:2411.00186}.
\newblock


\bibitem[Reimers and Gurevych(2019)]%
        {reimers2019sbert}
\bibfield{author}{\bibinfo{person}{Nils Reimers} {and} \bibinfo{person}{Iryna
  Gurevych}.} \bibinfo{year}{2019}\natexlab{}.
\newblock \showarticletitle{Sentence-{BERT}: Sentence Embeddings using Siamese
  {BERT}-Networks}.
\newblock \bibinfo{journal}{\emph{Proceedings of EMNLP-IJCNLP}}
  (\bibinfo{year}{2019}).
\newblock


\bibitem[{\c{S}}enol et~al\mbox{.}(2026)]%
        {senol2025dkllm}
\bibfield{author}{\bibinfo{person}{Ali {\c{S}}enol}, \bibinfo{person}{Garima
  Agrawal}, {and} \bibinfo{person}{Huan Liu}.} \bibinfo{year}{2026}\natexlab{}.
\newblock \showarticletitle{Domain Knowledge-Enhanced {LLMs} for Fraud and
  Concept Drift Detection}.
\newblock \bibinfo{journal}{\emph{Electronics}} \bibinfo{volume}{15},
  \bibinfo{number}{3} (\bibinfo{year}{2026}), \bibinfo{pages}{534}.
\newblock
\shownote{arXiv:2506.21443}.
\newblock


\bibitem[{SNAP}(2019)]%
        {mooc2019snap}
\bibfield{author}{\bibinfo{person}{{SNAP}}.} \bibinfo{year}{2019}\natexlab{}.
\newblock \bibinfo{title}{{MOOC} User Action Dataset}.
\newblock
\shownote{Stanford Network Analysis Project}.
\newblock


\bibitem[Spiliopoulou et~al\mbox{.}(2006)]%
        {spiliopoulou2006monic}
\bibfield{author}{\bibinfo{person}{Myra Spiliopoulou}, \bibinfo{person}{Irene
  Ntoutsi}, \bibinfo{person}{Yannis Theodoridis}, {and} \bibinfo{person}{Rene
  Schult}.} \bibinfo{year}{2006}\natexlab{}.
\newblock \showarticletitle{{MONIC}: Modeling and Monitoring Cluster
  Transitions}. In \bibinfo{booktitle}{\emph{Proceedings of the 12th ACM SIGKDD
  International Conference on Knowledge Discovery and Data Mining (KDD)}}.
\newblock


\bibitem[Viswanathan et~al\mbox{.}(2024)]%
        {viswanathan2024fewshot}
\bibfield{author}{\bibinfo{person}{Vijay Viswanathan}, \bibinfo{person}{Kiril
  Gashteovski}, \bibinfo{person}{Carolin Lawrence}, \bibinfo{person}{Tongshuang
  Wu}, {and} \bibinfo{person}{Graham Neubig}.} \bibinfo{year}{2024}\natexlab{}.
\newblock \showarticletitle{Large Language Models Enable Few-Shot Clustering}.
\newblock \bibinfo{journal}{\emph{Transactions of the Association for
  Computational Linguistics (TACL)}} (\bibinfo{year}{2024}).
\newblock
\shownote{arXiv:2307.00524}.
\newblock


\bibitem[Wan et~al\mbox{.}(2024)]%
        {wan2024tntllm}
\bibfield{author}{\bibinfo{person}{Mengting Wan}, \bibinfo{person}{Tara
  Safavi}, \bibinfo{person}{Sujay~Kumar Jauhar}, \bibinfo{person}{Yujin Kim},
  \bibinfo{person}{Scott Counts}, \bibinfo{person}{Jennifer Neville},
  \bibinfo{person}{Siddharth Suri}, \bibinfo{person}{Chirag Shah},
  \bibinfo{person}{Ryen~W. White}, \bibinfo{person}{Longqi Yang},
  \bibinfo{person}{Reid Andersen}, \bibinfo{person}{Georg Buscher},
  \bibinfo{person}{Dhruv Joshi}, {and} \bibinfo{person}{Nagu Rangan}.}
  \bibinfo{year}{2024}\natexlab{}.
\newblock \bibinfo{title}{{TnT-LLM}: Text Mining at Scale with Large Language
  Models}.
\newblock
\shownote{arXiv:2403.12173}.
\newblock


\bibitem[{YOOCHOOSE GmbH}(2015)]%
        {yoochoose2015}
\bibfield{author}{\bibinfo{person}{{YOOCHOOSE GmbH}}.}
  \bibinfo{year}{2015}\natexlab{}.
\newblock \bibinfo{title}{{RecSys} Challenge 2015 ({YOOCHOOSE}) Dataset}.
\newblock
\shownote{ACM RecSys Challenge 2015}.
\newblock


\bibitem[Zhang et~al\mbox{.}(2023)]%
        {zhang2023clusterllm}
\bibfield{author}{\bibinfo{person}{Yuwei Zhang}, \bibinfo{person}{Zihan Wang},
  {and} \bibinfo{person}{Jingbo Shang}.} \bibinfo{year}{2023}\natexlab{}.
\newblock \showarticletitle{{ClusterLLM}: Large Language Models as a Guide for
  Text Clustering}. In \bibinfo{booktitle}{\emph{Proceedings of EMNLP}}.
\newblock
\shownote{arXiv:2305.14871; verified}.
\newblock


\end{thebibliography}

\appendix
\onecolumn

\section{Appendix overview}
This appendix documents the method in enough detail to audit the evaluation and the implementation choices
without the source code. It gives the exact LLM prompt (\S\ref{app:prompt}), the precise cluster-state
\emph{artifact} passed to the model and a real example on public data (\S\ref{app:artifact}), the arbiter's
decisions on that example (\S\ref{app:example}), the drift-injection protocols (\S\ref{app:injection}),
feature construction (\S\ref{app:features}), baseline threshold settings (\S\ref{app:baselines}), and all
hyperparameters and evaluation constants (\S\ref{app:hyper}). It then reports robustness analyses and
ablations (\S\ref{app:robustness}), the window-tolerance sensitivity (\S\ref{sec:supp-tol}), and additional
result figures (\S\ref{sec:supp-figs}). Every worked example is drawn from the \textbf{MOOC} public dataset.

\section{Arbiter prompt}\label{app:prompt}
The arbiter is a single prompt with a \texttt{\{cluster\_report\}} slot. The three ablation channels use
structurally identical prompts that differ only in which report content they receive (and a one-line
preamble): \textbf{combined} (below), \textbf{stats-only} (identical but the report omits labels and member
texts, and the ``use both channels'' principle is dropped), and \textbf{semantics-only} (the report omits
all numeric health signals, so the model sees only labels and representative member texts). The prompt
contains \emph{no numeric cutoffs}. The model is instructed to judge relative to each cluster's own history
and the rest of the population.

\begin{lstlisting}
You are a cluster lifecycle analyst. You monitor behavioral session clusters and decide
whether each cluster needs a structural change.

## Context
You manage a clustering system that groups user sessions (sequences of actions) into
behavioral clusters. Each cluster was created to capture ONE coherent behavior, described by
its label and representative journeys. New data arrives each time window. Over time a cluster
can drift away from the behavior it was built for. Your job is to judge -- from each cluster's
current state and what it was meant to represent -- whether its structure should change.

## Cluster Report
EVERY active cluster this window is listed below (this is NOT a pre-filtered subset). Each
comes with its original behavioral identity (label + representative member texts), its recent
quantitative health signals and how they have moved over the windows, plus a population-spread
summary and a residual-sessions section. Most clusters in most windows need NO_ACTION -- judge
each RELATIVE to the others shown and to its own history.

{cluster_report}

## How to reason
Do not apply fixed numeric cutoffs. Instead, weigh the evidence holistically: compare each
signal to that cluster's OWN history and to the other clusters this window, and ask whether the
change is large and sustained enough to be a real structural shift rather than normal
fluctuation. Behavioral data is noisy -- a single window's movement, or a change comparable to
what other stable clusters show, is usually just noise. Let the behavioral description anchor
your judgment: a statistical wobble that doesn't correspond to a plausible change in what users
are actually doing is probably not actionable.

## Available Actions -- what each one means
- NO_ACTION: The cluster still coherently represents its original behavior. Its signals are
  within the range of normal variation for this data. Correct for most clusters, most windows.
- SPLIT: The cluster now contains two genuinely distinct behaviors that were one at creation.
  Evidence: cohesion has degraded meaningfully and persistently (not a one-window blip), members
  appear to separate into sub-groups, and the behavioral pattern suggests two intents conflated.
- MERGE: A shrinking cluster's members are migrating into a specific, highly-similar sibling --
  the two now represent essentially the same behavior. Report MERGE on the shrinking cluster and
  name the sibling as merge_with.
- RETIRE: The cluster's behavior is no longer occurring -- members have dwindled toward zero and
  dispersed (not migrated to one sibling). The behavior it captured has stopped.
- CREATE: A coherent new behavior has appeared that no existing cluster represents -- the
  residual-sessions section shows a sizeable group of sessions far from every current centroid
  whose texts tell one consistent NEW story. Use cluster_id "NEW".

## Judgment principles
- Prefer NO_ACTION when uncertain. Unnecessary interventions are more costly than a delayed one,
  and over-splitting is the most common failure.
- Use both channels together, but do not require both to be strong. A clear, sustained signal in
  EITHER channel can justify action as long as the other channel does not actively contradict it.
  Treat the two channels as complementary evidence, not as an AND-gate. Reject a signal that one
  channel asserts and the other clearly refutes.
- Judge each signal against this cluster's own baseline and the spread of the other clusters --
  "high" only means high relative to what is normal here.

## Output Format
Respond with a JSON array, one entry per cluster in the report (or "NEW" for CREATE). In each
`reasoning`, briefly state what the state + behavior told you and why it does or doesn't warrant
action:
[
  {"cluster_id": 0, "action": "NO_ACTION", "reasoning": "..."},
  {"cluster_id": 1, "action": "SPLIT", "reasoning": "..."},
  {"cluster_id": "NEW", "action": "CREATE", "reasoning": "..."}
]
Only include clusters that appear in the report. Return ONLY the JSON array, no other text.
\end{lstlisting}

\section{The cluster-state artifact passed to the LLM}\label{app:artifact}
Each window the \texttt{\{cluster\_report\}} slot is filled with a plain-text report built deterministically
from the registry. Its schema: (i) a one-line \emph{window state} (active-cluster count, arrivals);
(ii) a \emph{population-spread summary} giving the min/median/max of cohesion-$z$, membership, and
nearest-neighbor similarity across clusters (so every signal can be read relative to the current
distribution); (iii) a \emph{residual-sessions} block (count and internal-spread ratio of items beyond a
scale-free mean$+1.5\sigma$ distance cut, the \textsc{create} candidate), and (iv) one block \emph{per
active cluster}: its label, member count, the last-5-window cohesion history, cohesion-$z$ (vs.\ the
cluster's own history), nearest-cluster similarity, the $k{=}2$ split-improvement ratio, and
distance-to-centroid mean/std. The \textbf{combined} channel includes labels and (in the per-cluster and
residual blocks) representative member texts. \textbf{Stats-only} omits them, and \textbf{semantics-only}
omits the numeric lines. Figure text below is a \emph{real, unedited} report (MOOC, seed 42, window 6).

\begin{lstlisting}
## Window state: 8 active clusters, 3149 sessions arrived this window.
Every active cluster is listed below (this is NOT a pre-filtered subset). Most clusters in most
windows should be NO_ACTION. Judge each cluster RELATIVE to the others here and to its own
history -- 'high'/'low'/'shrinking' mean so only compared with this population.

- Population spread -- cohesion Z: min=1.80 median=2.83 max=10.30; members: min=30.00
  median=473.50 max=606.00; nearest-neighbor similarity: min=0.81 median=0.94 max=0.95

### Residual sessions -- poorly explained by ANY current cluster (candidate new behavior)
- Count beyond mean+1.5*std of distance-to-nearest-centroid: 190 (6.0% of this window)
- Residual group internal spread: 0.875 (vs avg cluster spread 0.437; ratio 2.00 -- lower ratio
  = residuals form one tight group rather than scattered noise)

### Cluster 0: "Progressing through structured learning content with repetitive practice"
- Members this window: 357
- Cohesion history (last 5): ['0.214', '0.302', '0.343', '0.392', '0.588']
- Cohesion Z-score: 4.45
- Nearest cluster: C9 "Browsing content then engaging with learning activities" (sim: 0.904)
- Split improvement ratio: 0.568 (mean sub-cluster cohesion / parent cohesion; lower = more sub-structure)
- Distance to centroid: mean=0.588, std=0.607

### Cluster 1: "Exploring content with repetitive engagement and backtracking patterns"
- Members this window: 439
- Cohesion history (last 5): ['0.208', '0.264', '0.279', '0.280', '0.319']
- Cohesion Z-score: 2.26
- Nearest cluster: C7 "Exploring content with repeated engagement and navigation patterns" (sim: 0.946)
- Split improvement ratio: 0.755
- Distance to centroid: mean=0.319, std=0.065

### Cluster 2: "Browsing content with frequent check-ins and returns"
- Members this window: 30
- Cohesion history (last 5): ['0.175', '0.184', '0.261', '0.397', '0.396']
- Cohesion Z-score: 2.07
- Nearest cluster: C5 "Exploring content with repetitive engagement patterns" (sim: 0.953)
- Split improvement ratio: 0.961
- Distance to centroid: mean=0.396, std=0.010

### Cluster 4: "Progressing through structured learning content with repeated practice"
- Members this window: 583
- Cohesion history (last 5): ['0.201', '0.365', '0.333', '0.333', '0.521']
- Cohesion Z-score: 3.40
- Nearest cluster: C0 "Progressing through structured learning content with repetitive practice" (sim: 0.813)
- Split improvement ratio: 0.695
- Distance to centroid: mean=0.521, std=0.303

### Cluster 5: "Exploring content with repetitive engagement patterns"
- Members this window: 66
- Cohesion history (last 5): ['0.224', '0.225', '0.234', '0.242', '0.263']
- Cohesion Z-score: 3.99
- Nearest cluster: C2 "Browsing content with frequent check-ins and returns" (sim: 0.953)
- Split improvement ratio: 0.844
- Distance to centroid: mean=0.263, std=0.079

### Cluster 6: "Focused practice with repeated activity engagement patterns"
- Members this window: 606
- Cohesion history (last 5): ['0.218', '0.276', '0.343', '0.344', '0.369']
- Cohesion Z-score: 1.80
- Nearest cluster: C5 "Exploring content with repetitive engagement patterns" (sim: 0.943)
- Split improvement ratio: 0.854
- Distance to centroid: mean=0.369, std=0.054

### Cluster 7: "Exploring content with repeated engagement and navigation patterns"
- Members this window: 560
- Cohesion history (last 5): ['0.210', '0.289', '0.286', '0.300', '0.325']
- Cohesion Z-score: 1.84
- Nearest cluster: C1 "Exploring content with repetitive engagement and backtracking patterns" (sim: 0.946)
- Split improvement ratio: 0.762
- Distance to centroid: mean=0.325, std=0.065

### Cluster 9: "Browsing content then engaging with learning activities"
- Members this window: 508
- Cohesion history (last 5): ['0.214', '0.265', '0.297', '0.325', '0.716']
- Cohesion Z-score: 10.30
- Nearest cluster: C0 "Progressing through structured learning content with repetitive practice" (sim: 0.904)
- Split improvement ratio: 0.475
- Distance to centroid: mean=0.716, std=0.742
\end{lstlisting}

\section{Arbiter decisions on the example}\label{app:example}
Given the report above, the combined arbiter returned the following (unedited) decisions for the acting
clusters. The remaining five clusters received \textsc{no\_action} with per-cluster rationale (omitted for
space). Note the model's reliance on \emph{sustained} degradation and split-improvement ratio, and its
explicit caution on the marginal \textsc{create}.

\begin{lstlisting}
{"cluster_id": 0, "action": "SPLIT", "reasoning": "Cohesion has degraded sharply and
 persistently over the last 5 windows (0.214->0.588, Z=4.45), distance-to-centroid std is very
 high (0.607), and the split improvement ratio (0.568) is among the lowest here, indicating
 strong sub-structure. The 'structured learning with repetitive practice' label plausibly
 conflates distinct progression vs. practice intents. Sustained degradation, not a blip."}

{"cluster_id": 9, "action": "SPLIT", "reasoning": "Extreme, sustained cohesion degradation
 (0.214->0.716 over 5 windows, Z=10.30, highest here), very high distance std (0.742), and lowest
 split improvement ratio (0.475) -- strongest sub-structure signal in the population. The
 'browsing then engaging' label spans two sequential behaviors that are plausibly splitting
 apart. Clear structural break."}

{"cluster_id": "NEW", "action": "CREATE", "reasoning": "Residual group is sizeable (190, 6% of
 window) and its internal spread ratio (2.00) -- while showing they're not one perfectly tight
 cluster -- combined with the count suggests an emerging coherent behavior poorly explained by
 any centroid. Marginal, but worth surfacing as a candidate new behavior."}
\end{lstlisting}

\section{Drift-injection protocols}\label{app:injection}
Ground-truth events are code-injected so decisions can be scored, and the base data is otherwise untouched.
Each seed injects the same 7 events (2 \textsc{split}, 2 \textsc{merge}, 2 \textsc{retire}, 1
\textsc{create}), and the remaining clusters are \textsc{no\_action} controls.

\paragraph{MOOC and YooChoose (vector perturbation).} On the 387-d composite features:
\begin{itemize}[leftmargin=1.4em,itemsep=1pt]
  \item \textbf{\textsc{split}} (clusters $0,4$; start windows $3,7$; duration 3): a fraction ($50\%$, then
        $70\%$) of the cluster's members are shifted progressively along a fixed random direction with
        magnitude $s\cdot\text{prog}\cdot\lVert\text{centroid}\rVert$, $s=0.60$ (the
        \texttt{split\_magnitude\_scale}; raised from $0.15$ because the resulting cohesion signal
        otherwise sat below natural window-to-window noise).
  \item \textbf{\textsc{merge}} (pairs $(2,7),(1,5)$; start $5,8$): the absorbed cluster's members move
        toward the sibling centroid with strength ramping $0.4\!\to\!0.9$ over three windows (absorption
        model; the survivor grows, the absorbed shrinks).
  \item \textbf{\textsc{retire}} (clusters $8,9$; start $4,7$): members are reassigned to their nearest
        \emph{other} centroid (scatter to zero).
  \item \textbf{\textsc{create}} (start window 6): a novel center is placed at $2.5\times$ the average
        centroid norm (maximizing min-distance to existing centroids over 20 draws); a growing fraction
        ($5\%\!\to\!18\%$) of each window's items is moved there. A boolean \emph{create-population mask}
        records exactly which items are the injected novelty, for strict \textsc{create} scoring.
\end{itemize}

\paragraph{News-Category (label-composition curation).} Real MiniLM embeddings are \emph{never}
perturbed; drift is induced by curating which real articles appear in which window:
\begin{itemize}[leftmargin=1.4em,itemsep=1pt]
  \item \textbf{\textsc{split}}: within-category along a category's own $k{=}2$ axis (e.g.\ a wellness
        category into mental- vs.\ physical-health sub-themes); the emerging sub-population ramps in from
        the split window at its natural $\sim$25\% minority share (not gamed to $50/50$). This is a
        \emph{semantic} split by construction: the geometric $k{=}2$ split ratio stays $\sim$0.98
        (stats-only cannot see it) while the member texts shift from one theme to two. All three LLM modes
        score $0$ \textsc{split} F1 on News, and the threshold and refit baselines register non-zero
        \textsc{split} F1 there only by firing on most clusters in most windows (Statistical Triggers: 23
        true positives against 318 false positives; MONIC: 5 against 20), so the event is undetected at
        usable precision by any method, not an honest null for all.
  \item \textbf{\textsc{merge}}: real near-duplicate category pairs; the absorbed category ramps to zero as
        the survivor grows (member conservation).
  \item \textbf{\textsc{retire}}: a low-volume category ramps to zero; members disperse.
  \item \textbf{\textsc{create}}: a held-out category is withheld initially and introduced mid-stream; its
        articles are the create-population for strict scoring.
\end{itemize}

\section{Feature construction and preprocessing}\label{app:features}
Sessions/items are embedded with MiniLM (all-MiniLM-L6-v2, 384-d) \cite{reimers2019sbert}. MOOC and
YooChoose use a 387-d composite: L2 of the text embedding weighted $0.8$, concatenated with
StandardScaler-normalized $[\#\text{actions},\#\text{unique actions},\text{duration}]$ weighted $0.2$
(un-normalized Euclidean space). News uses the L2-normalized 384-d embedding directly (bounded cosine
space). Base clusterings: $K{=}10$ (MOOC, YooChoose) and $K{=}15$ (News), fit once at window 0. Windows:
12 equal-time windows per dataset. \emph{Label richness} $r$ is the mean pairwise cosine distinctiveness
($1-\text{cosine}$) of the cluster-label embeddings: $0.17$ (YooChoose), $0.50$ (MOOC), $0.695$ (News,
rounded to $0.70$ in the main text).

\section{Baseline settings}\label{app:baselines}
\begin{itemize}[leftmargin=1.4em,itemsep=1pt]
  \item \textbf{Statistical Triggers}: \textsc{split} if cohesion-$z\ge2.0$; \textsc{merge} if centroid
        cosine $>0.92$; \textsc{retire} if empty for $\ge2$ windows; \textsc{create} if orphan share
        $>10\%$ at $>2\times$ mean distance.
  \item \textbf{Smart Triggers} (tuned heuristic; \emph{not} a clean minus-LLM ablation): \textsc{split}
        if $z\ge2.0$ \emph{and} 3-window worsening \emph{and} $k{=}2$ ratio $<0.80$; \textsc{merge} if
        live similarity $>0.92$, converging over 3 windows, and one side shrinking; \textsc{retire} if
        member count $<5$ for $\ge2$ windows with no sibling $>0.88$; \textsc{create} if orphan share
        $>8\%$ with orphan-cohesion ratio $<1.5$. Pre-filter $z>1.8$/members$<5$/sim$>0.88$; hard split
        guard (block if $k{=}2$ ratio $>0.85$); cap 3 actions/window. These constants are calibrated on
        MOOC and do not transfer (F1 $0.272$ MOOC $\to 0.000$ News).
  \item \textbf{MONIC} (refit-and-track steelman): per-window $k$ chosen by silhouette over
        $k_{\text{prev}}\pm2$; align new$\leftrightarrow$previous clusters by centroid cosine with a floor
        $\tau$ (persistent ids seeded from window 0); transitions read from the bipartite match. The
        floor is \emph{per-dataset}: $\tau=0.97$ (YooChoose), $0.86$ (MOOC), $0.80$ (News), calibrated
        from stable cross-window self-similarity vs.\ distinct-cluster similarity. That MONIC needs
        per-dataset $\tau$ while the arbiter uses one prompt is part of the transfer argument.
\end{itemize}

\section{Hyperparameters and evaluation constants}\label{app:hyper}
\begin{itemize}[leftmargin=1.4em,itemsep=1pt]
  \item \textbf{Seeds (15)}: 42, 123, 456, 789, 1024, 2048, 3000, 4096, 5555, 7777, 8191, 9001, 11111,
        13337, 24601.
  \item \textbf{Arbiter LLM}: Claude Opus~4.8 (\texttt{us.anthropic.claude-opus-4-8}); this model exposes no
        temperature control, so the arbiter is near-deterministic; $\sim$1 call per window ($\sim$10 calls/seed/mode). The
        \emph{same prompt} across all datasets. Per-token usage was not captured because the inference client
        did not populate it, so the cost-accounting limitation reports a chars/4 token estimate instead.
  \item \textbf{Residual surfacing}: items beyond mean$+1.5\sigma$ of distance-to-nearest-centroid
        (scale-free); \textsc{create} execution promotes the orphan mass above the 85th distance
        percentile. Skip-first-2-windows data floor. Hard split guard default \emph{off} for the arbiter.
  \item \textbf{Evaluation}: a decision is a true positive only within $\pm2$ windows of the ground-truth
        window; \textsc{split}/\textsc{retire} require the correct target; \textsc{merge} is
        direction-agnostic but requires both clusters of the pair to match (overlap $\ge2$);
        \textsc{create} requires create-population purity $\ge0.5$. Restraint $=$ fraction of control
        clusters never touched. Quality (silhouette) scored as-of-window from created/retired-window
        bookkeeping.
  \item \textbf{Significance}: one-sided paired Wilcoxon signed-rank on per-seed F1 (best LLM vs.\ best
        baseline) plus a $10{,}000$-sample paired bootstrap 95\% CI on the F1 delta.
\end{itemize}

\section{Robustness analyses and ablations}\label{app:robustness}
These additive analyses stress-test the main claims. None changes the cemented results. Each reuses the
saved per-window decision logs and the same strict scorer.

\paragraph{Label-causality ablation.} We re-run the semantic modes with the cluster labels (and, on News,
the member texts) either \emph{shuffled} across clusters or replaced by \emph{generic} placeholders, to test
whether the richness pattern is causal or only correlational. Table~\ref{tab:ablation} shows it is
causal. On News (rich labels) destroying the labels collapses \textbf{combined} from $0.228$ to $0.000$, the
stats-only floor, so the semantic channel is causally necessary there. On YooChoose (impoverished labels)
the score does not fall, and \textbf{semantics-only} in fact rises from $0.083$ to $0.393$ once the
uninformative labels are neutralized, showing those labels were mildly harmful. This
establishes the necessity and harm directions of the richness pattern. We do not test sufficiency, which
would require synthesizing richer labels.

\begin{table}[t]\centering\small
\caption{Label-causality ablation. F1 (mean over 15 seeds) with real, shuffled, and generic labels.}
\label{tab:ablation}
\begin{tabular}{llccc}
\toprule
Dataset & Mode & Real & Shuffle & Generic \\
\midrule
News & combined       & $0.228$ & $0.066$ & $0.000$ \\
News & semantics-only & $0.155$ & $0.091$ & $0.000$ \\
YooChoose & combined       & $0.176$ & $0.197$ & $0.249$ \\
YooChoose & semantics-only & $0.083$ & $0.082$ & $0.393$ \\
\bottomrule
\end{tabular}
\end{table}

\paragraph{Same-numeric-feature non-LLM arbiter.} To test whether the advantage comes from the report's
numeric signals or from the reasoning, we train a gradient-boosted classifier on the exact numeric
signals the arbiter sees (this baseline does not consume the semantic labels or member texts) and score it
with the same strict scorer (Table~\ref{tab:samefeature}). Even given per-dataset
supervision through leave-one-seed-out, which the LLM never receives, it does not exceed the LLM at usable
precision on any dataset. Its lone numeric edge (News LOSO $0.242$ vs $0.228$) is within seed noise and is a
low-precision over-fire. In the cross-dataset regime that matches the LLM's actual zero-per-dataset-tuning
setting it collapses to $0.06$ to $0.15$. The rich feature report alone does not explain the result.

\begin{table}[t]\centering\small
\caption{Same-numeric-feature learned arbiter (F1, 15 seeds). LOSO grants per-dataset supervision the LLM never
receives. Cross-dataset matches the LLM's zero-tuning transfer setting.}\label{tab:samefeature}
\begin{tabular}{lccc}
\toprule
Dataset & LOSO (per-dataset) & Cross-dataset & LLM \\
\midrule
MOOC      & $0.297$ & $0.064$ & $0.348$ \\
News      & $0.242$ & $0.146$ & $0.228$ \\
YooChoose & $0.308$ & $0.062$ & $0.336$ \\
\bottomrule
\end{tabular}
\end{table}

\paragraph{Oracle-tuned triggers (upper bound, not a baseline).} To address whether per-dataset threshold
tuning would close the gap, we grant the trigger baselines an \emph{oracle}. Their thresholds are
grid-searched against the evaluation ground truth, information a real streaming deployment never has, so
this is an unattainable upper bound, not a deployable competitor (Table~\ref{tab:oracle}).
Oracle-tuned statistical triggers reach only $0.09$ to $0.13$. Oracle-tuned \emph{smart} triggers tuned and
scored on the same seeds exceed the LLM's F1 on the numeric-drift datasets (MOOC, YooChoose) but score
\emph{zero} on the semantic-drift dataset (News), where an exhaustive search confirms no threshold setting
detects the event. Removing that overfit with held-out tuning (tune on train seeds, evaluate on disjoint
held-out seeds) leaves the smart-trigger upper bound only within-noise of the LLM on the numeric datasets
($0.381$ vs $0.348$, $0.356$ vs $0.336$), reached with a per-dataset labeled tuning budget the LLM does not
spend, and still zero on News. So no trigger, even oracle-tuned, detects semantic drift, and on numeric
drift the LLM is on par with a cheating upper bound while using no per-dataset tuning. Against the
\emph{deployable} baselines (no ground-truth access) the LLM leads on F1 on all three datasets.

\begin{table}[t]\centering\small
\caption{Trigger upper bounds vs the LLM (F1, 15 seeds). The oracle columns select thresholds using the
evaluation ground truth and are unattainable upper bounds, not deployable methods.}\label{tab:oracle}
\begin{tabular}{lccccc}
\toprule
 & Untuned & Oracle & Oracle smart & Held-out & LLM \\
Dataset & smart & stat & (overfit) & oracle smart & \\
\midrule
MOOC      & $0.272$ & $0.133$ & $0.406$ & $0.381$ & $0.348$ \\
News      & $0.000$ & $0.087$ & $0.000$ & $0.000$ & $0.228$ \\
YooChoose & $0.223$ & $0.105$ & $0.401$ & $0.356$ & $0.336$ \\
\bottomrule
\end{tabular}
\end{table}

\paragraph{Signal parity.} The report shows current membership but not a membership \emph{trend}, a signal
the smart triggers use for \textsc{merge} and \textsc{retire}. Surfacing a raw membership-history line to the
arbiter and re-running MOOC and YooChoose does not improve it (F1 unchanged to slightly lower on the winning
modes), so the on-par numeric result is not an information-asymmetry artifact.

\section{Window-tolerance sensitivity}\label{sec:supp-tol}
The main text credits a decision as a true positive when it lands within $\pm2$ windows of the
injected event. This tolerance reflects that drift is injected as a multi-window ramp rather than at a
single instant. On MOOC, for example, a \textsc{split} completes at its nominal window only after members
migrate over the preceding two to three windows (a ground-truth \textsc{split} at window~5 is injected as
``50\% of members split over W3--5''), so a correct detection can legitimately fire anywhere along that
ramp. To show the headline is not an artifact of this choice, we re-score the \emph{same} saved decision
logs at $\pm1$, $\pm2$, and $\pm3$ windows. This changes only the matcher, not the arbiter, so it involves
no additional LLM calls.

\begin{table}[t]\centering
\small
\caption{Best-LLM vs.\ best-baseline decision F1 (mean over 15 seeds) as the matching tolerance varies.
The winning method is unchanged on News and YooChoose at every tolerance. On MOOC the LLM wins at $\pm2$
and $\pm3$, while Smart-Triggers edges it at the tightest $\pm1$, where the LLM's ramp-latent detections
are clipped most.}\label{tab:tol}
\begin{tabular}{llccc}
\toprule
Dataset & Contrast & $\pm1$ & $\pm2$ & $\pm3$ \\
\midrule
\multirow{2}{*}{MOOC}
 & LLM (stats-only)     & $0.192$ & $\mathbf{0.348}$ & $\mathbf{0.348}$ \\
 & Best baseline (Smart-Triggers) & $\mathbf{0.252}$ & $0.272$ & $0.272$ \\
\midrule
\multirow{2}{*}{News}
 & LLM (combined)       & $\mathbf{0.173}$ & $\mathbf{0.228}$ & $\mathbf{0.257}$ \\
 & Best baseline (MONIC) & $0.075$ & $0.116$ & $0.143$ \\
\midrule
\multirow{2}{*}{YooChoose}
 & LLM (stats-only)     & $\mathbf{0.259}$ & $\mathbf{0.336}$ & $\mathbf{0.336}$ \\
 & Best baseline (Smart-Triggers) & $0.119$ & $0.223$ & $0.223$ \\
\bottomrule
\end{tabular}
\end{table}

Table~\ref{tab:tol} shows the ordering is stable on News and YooChoose across all three tolerances, with
the LLM ahead throughout and the gap widening slightly as the window loosens. MOOC is the one reversal, and
we report it plainly. At $\pm1$ Smart-Triggers ($0.252$) edges the best LLM mode ($0.192$) because the
trigger fires punctually at the statistical peak while the arbiter often acts one window into the ramp. At
$\pm2$, which matches the width over which MOOC drift is actually injected, and at $\pm3$ the LLM is clearly
ahead ($0.348$ vs.\ $0.272$). We therefore read $\pm2$ as the tolerance that fits the benchmark's own
injection design rather than one chosen to favor a method. The tightest setting that would disadvantage the
LLM is $\pm1$, and even there the LLM still wins on two of the three datasets.

\section{Additional result figures}\label{sec:supp-figs}
\begin{figure*}[t]\centering
\includegraphics[width=\linewidth]{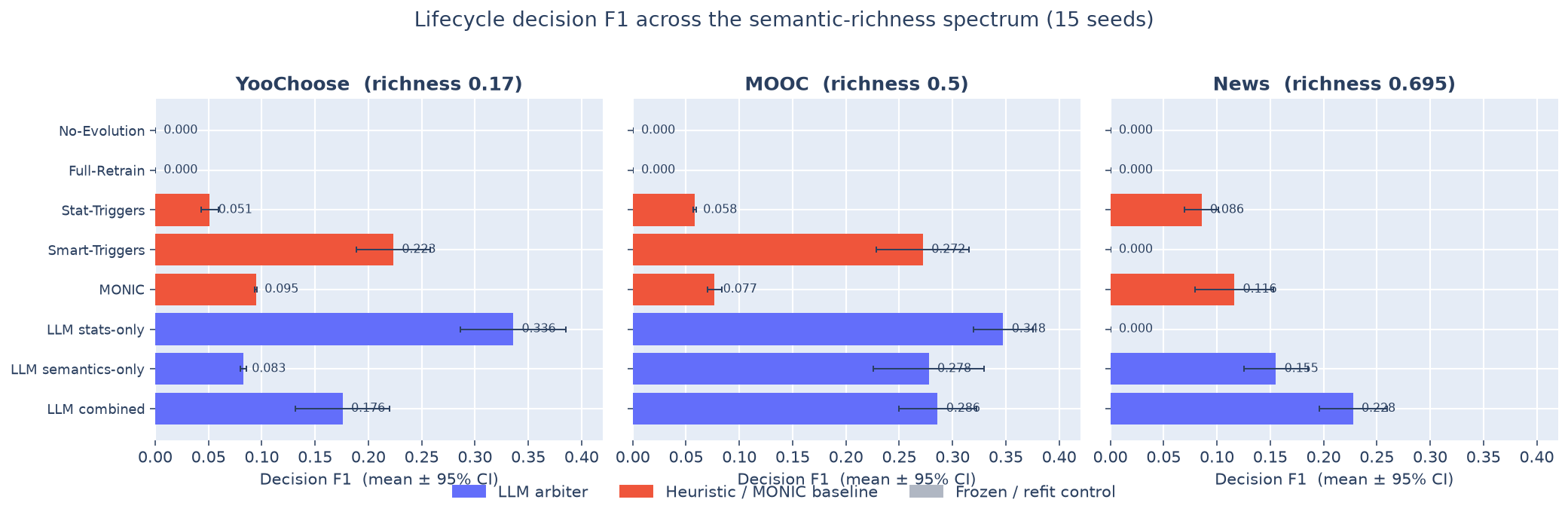}
\caption{Decision F1 by method on each dataset (15 seeds, $\pm$95\% CI). An LLM mode is best on every
dataset.}\Description{Grouped bar chart of decision F1 by method on each of the three datasets, with 95
percent confidence interval bars. An LLM mode has the highest F1 on every dataset.}\label{fig:mainf1}
\end{figure*}

\begin{figure}[t]\centering
\includegraphics[width=0.45\linewidth]{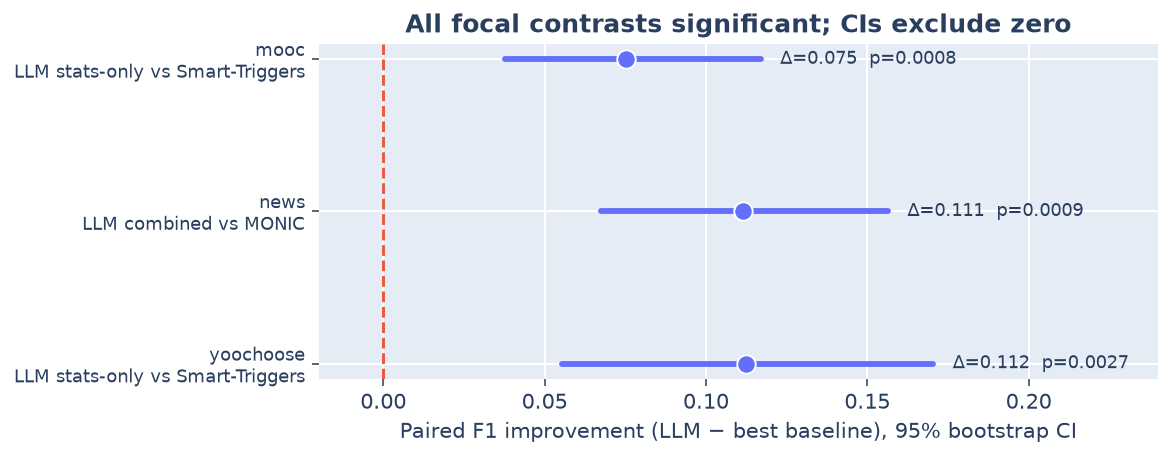}
\caption{Paired F1 improvement (LLM $-$ best baseline) with 95\% bootstrap CIs, all excluding zero.}
\Description{Bar chart of the paired F1 improvement of the best LLM mode over the best baseline on each
dataset, with 95 percent bootstrap confidence interval bars, all lying above zero.}
\label{fig:sig}
\end{figure}

\begin{figure*}[t]\centering
\includegraphics[width=\linewidth]{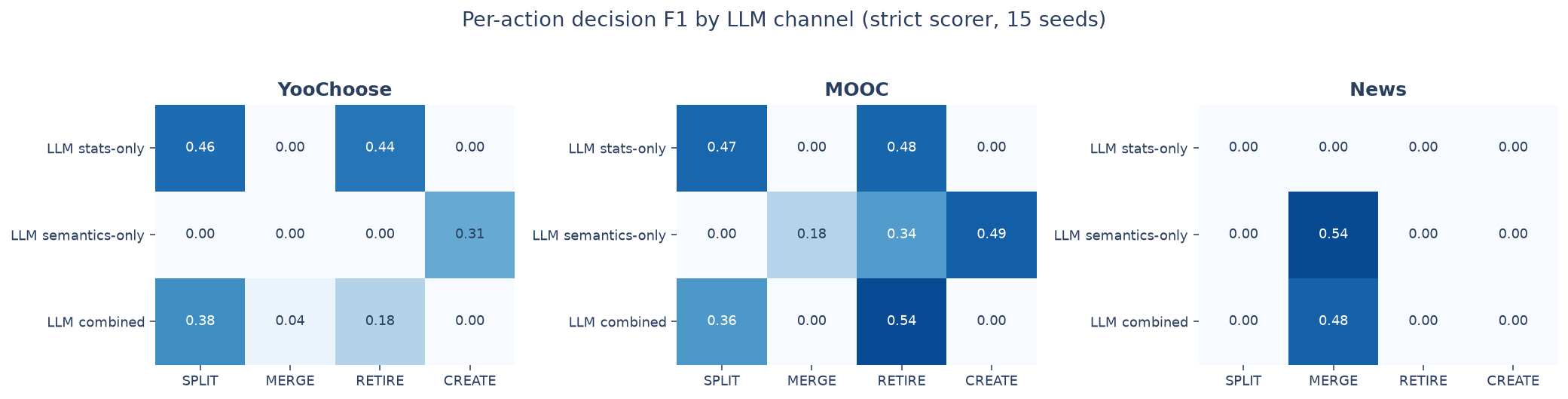}
\caption{Per-action decision F1 by LLM channel and dataset.}\Description{Grouped bar chart of per-action
decision F1 for split, merge, retire, and create, broken out by LLM channel and dataset, showing which
actions each channel detects on each dataset.}\label{fig:peraction}
\end{figure*}

\begin{figure*}[t]\centering
\includegraphics[width=\linewidth]{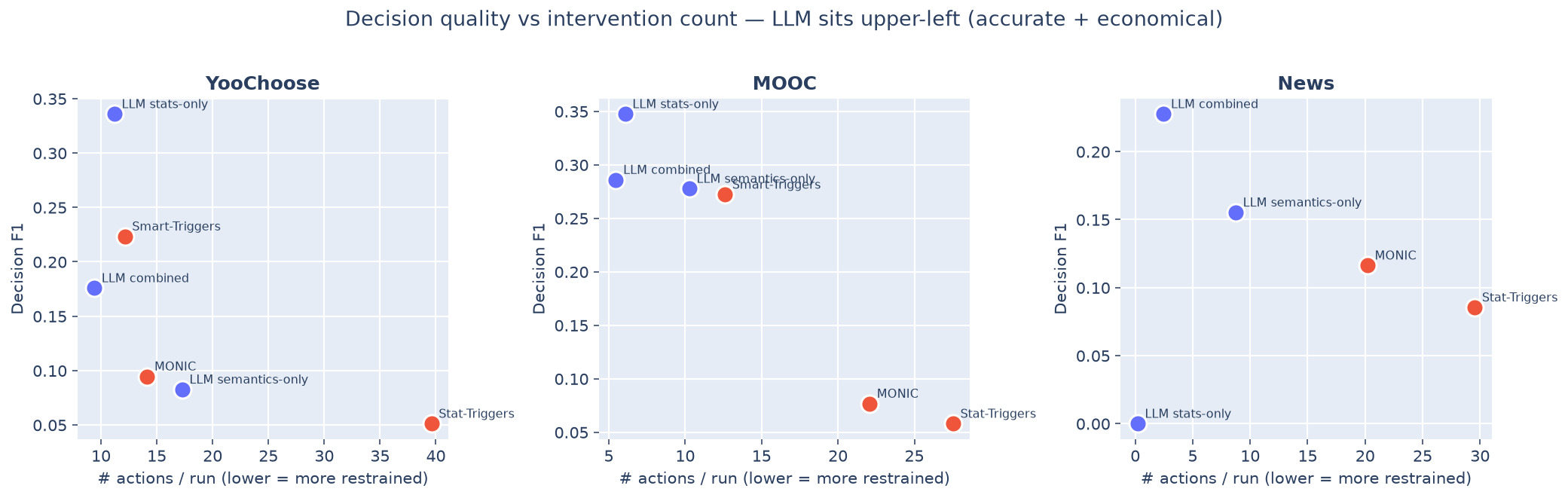}
\caption{Decision F1 vs.\ number of actions per run. LLM arbiter: accurate and economical.}\Description{Scatter
plot of decision F1 against the number of actions taken per run. The LLM arbiter points fall in the high-F1,
low-action region, indicating accurate and economical maintenance.}\label{fig:eff}
\end{figure*}

\begin{figure*}[t]\centering
\includegraphics[width=0.95\linewidth]{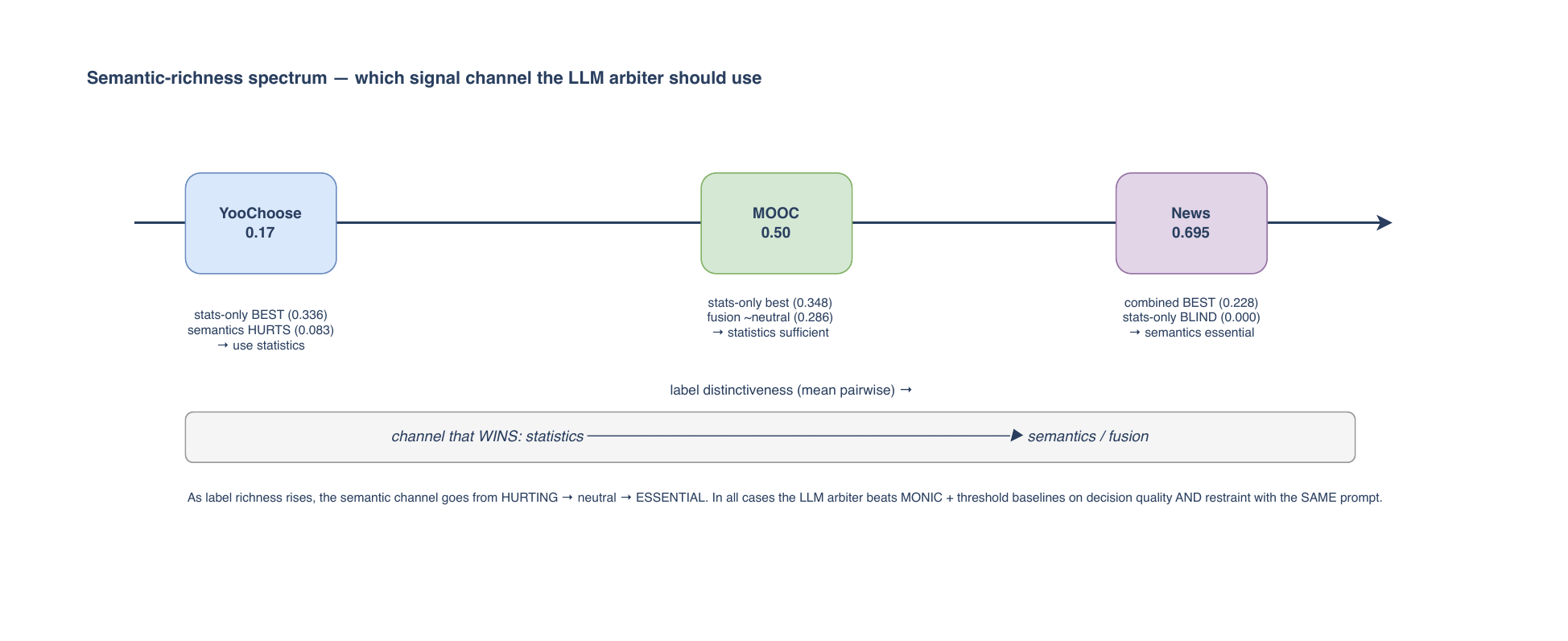}%
\caption{Conceptual spectrum: as label richness rises, the channel that wins moves
from statistics to semantics/fusion. The LLM arbiter beats mechanical baselines throughout with one
prompt.}\Description{Conceptual spectrum showing that as cluster-label richness increases, the
best-performing signal channel shifts from statistics-only to fused statistics-plus-semantics, while the LLM
arbiter stays ahead of the mechanical baselines throughout using a single
prompt.}\label{fig:spectrum-concept}\end{figure*}

\begin{figure*}[t]\centering
\includegraphics[width=0.95\linewidth]{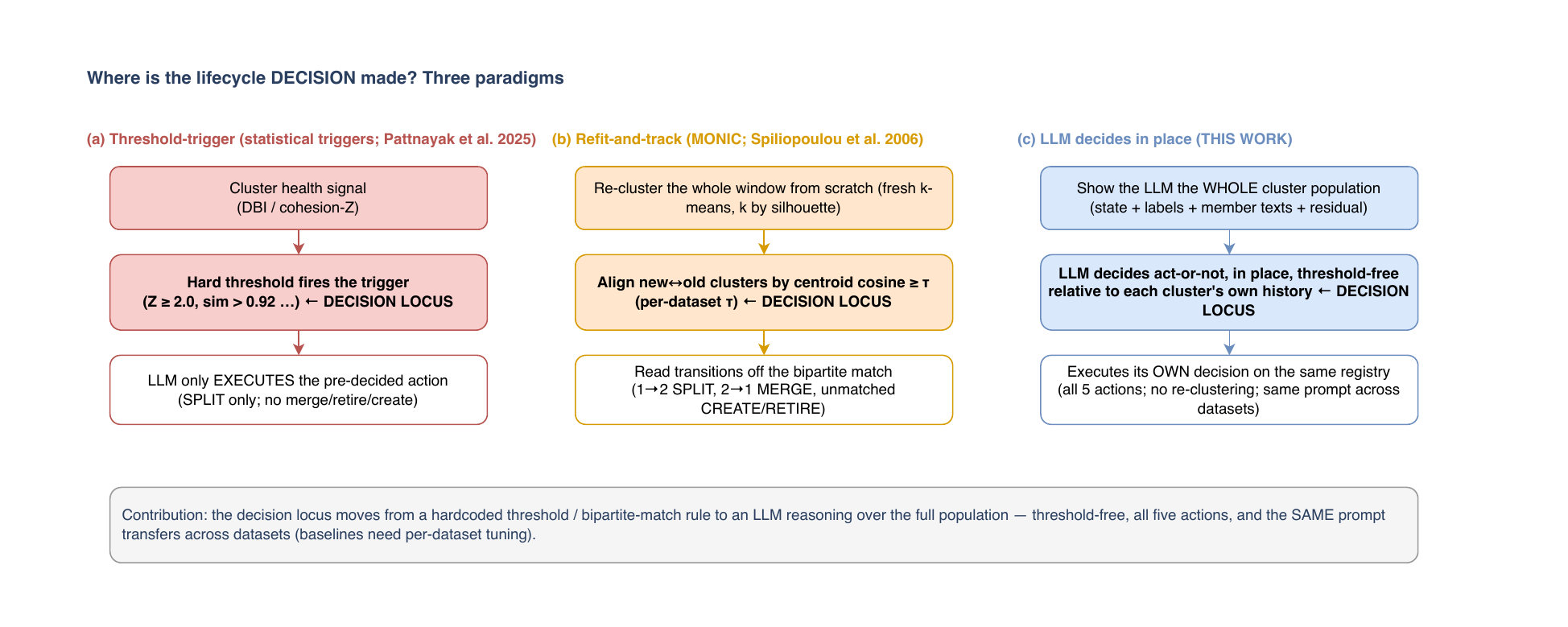}%
\caption{Three paradigms and where each makes the decision: (a) threshold
trigger, (b) MONIC refit-and-track, (c) LLM decides in place (this work).}\Description{Side-by-side
comparison of three lifecycle-maintenance paradigms and where each makes its decision: a threshold trigger
fires on a hardcoded statistic, MONIC re-clusters every window and reads transitions off the alignment, and
the proposed method places the decision in an LLM acting in place over the whole population with no
re-clustering.}\label{fig:paradigms}\end{figure*}

These figures re-plot numbers already tabulated in the main text. Figure~\ref{fig:mainf1} visualizes
Table~\ref{tab:main}, Figure~\ref{fig:sig} visualizes Table~\ref{tab:sig}, and Figure~\ref{fig:peraction}
visualizes Table~\ref{tab:peraction}. Figure~\ref{fig:eff} plots decision F1 against the number of actions per
run, and the two conceptual diagrams illustrate the semantic-richness spectrum (Figure~\ref{fig:spectrum-concept})
and the three lifecycle-decision paradigms (Figure~\ref{fig:paradigms}).

\end{document}